\documentclass[letterpaper, 10 pt, conference]{ieeeconf}  

\IEEEoverridecommandlockouts                              

\usepackage{graphicx}
\usepackage{hyperref}
\usepackage{xcolor} 

\hypersetup{
  colorlinks=true,
  linkcolor=blue,
  urlcolor=blue,
  citecolor=blue
}

\usepackage{amsfonts}
\usepackage{amssymb}

\DeclareMathAlphabet{\mathbb}{U}{msb}{m}{n}
\usepackage{algorithm}
\usepackage[noend]{algpseudocode}
\algrenewcommand\algorithmicindent{0.6em}%

\usepackage{mathalfa}
\usepackage{mathrsfs}
\usepackage{bbm}
\usepackage{amsmath}
\usepackage{float}
\usepackage{stmaryrd}
\usepackage{float} 
\usepackage{xspace}
\usepackage{rotating}
\usepackage{tikz} 
\usetikzlibrary{positioning}

\usepackage{amsmath,amssymb}
\usepackage{amssymb}
\usepackage{mathbbol}
\usepackage{tabto}
\usepackage{multirow}
\usepackage{graphicx}
\usepackage{array}
\usepackage{mathrsfs}
\usepackage{color}

\usepackage{pgfplots}
\usepgfplotslibrary{groupplots} 
\pgfplotsset{compat=1.18}

\usetikzlibrary{trees}
\DeclareMathOperator*{\argmax}{arg max} 

\usepackage[dvipsnames]{xcolor}
\usetikzlibrary{positioning, arrows.meta, fit, backgrounds, calc}

\makeatletter
\newcommand{\Break}{\STATE \@breaktrue \hfill\textbf{break}}
\newif\if@break
\makeatother

\DeclareMathAlphabet{\mathbb}{U}{msb}{m}{n}

\newcommand{\comment}[1]{}

\newcommand{\rob}{\ensuremath{\mathsf{R}}\xspace}
\newcommand{\taskR}{\ensuremath{\rob_{\textrm{task}}}\xspace}
\newcommand{\assistR}{\ensuremath{\rob_{\textrm{assist}}}\xspace}
\newcommand{\T}{\mathbb{T}}
\newcommand{\I}{\mathcal{I}}
\newcommand{\UB}{\mathcal{UB}}
\newcommand{\A}{\mathcal{A}}
\newcommand{\R}{\mathcal{R}}
\newcommand{\His}{\mathcal{H}}
\algnewcommand\cm[1]{\State // #1}
\algnewcommand\cmm[1]{\hfill // #1}

\newtheorem{dfn}{Definition}

\newtheorem{prob}{Problem}

\newcommand{\hists}{\mathcal{H}} 

\newcommand{\OS}[1]{\textcolor{red}{#1}}

\newcommand{\OD}[1]{\textcolor{blue}{#1}}

\newcommand{\ignore}[1]{}
\usepackage{subfig} 

\newcommand{\figPlacemnet}[2]{#2}

\usetikzlibrary{shapes.geometric, arrows.meta, positioning}
\usetikzlibrary{positioning, arrows.meta, fit, backgrounds, calc}
\usetikzlibrary{shapes.geometric, positioning}
\usetikzlibrary{backgrounds} 
\pgfdeclarelayer{background}
\pgfsetlayers{background,main}

\usepackage[english]{babel}

\usepackage{bbm}
\newtheorem{example}{Example}

\title{\LARGE \bf
Joint Task Assistance Planning via Nested Branch and Bound (Extended Version)
}

\author{
Omer Daube and Oren Salzman
\thanks{O. Daube and O. Salzman are with the Faculty of Computer Science,
        Technion, Haifa, Israel. E-mail: {\tt\small \{omer.daube, osalzman\}@cs.technion.ac.il}}%
}

\begin{document}

\maketitle
\thispagestyle{empty}
\pagestyle{empty}





\begin{abstract}
We introduce and study the Joint Task Assistance Planning problem which generalizes prior work on optimizing assistance in robotic collaboration. In this setting, two robots operate over predefined roadmaps, each represented as a graph corresponding to its configuration space. One robot, the task robot, must execute a timed mission, while the other, the assistance robot, provides sensor-based support that depends on their spatial relationship. The objective is to  compute a path for both robots that maximizes the total duration of assistance given. Solving this problem is challenging due to the combinatorial explosion of possible path combinations together with the temporal nature of the problem (time needs to be accounted for as well). To address this, we propose a nested branch-and-bound framework that efficiently explores the space of robot paths in a hierarchical manner. 
We empirically evaluate our algorithm and demonstrate a speedup of up to two orders of magnitude when compared to a baseline approach.
\end{abstract}

\section{Introduction}
\label{sec:intro}

In Task Assistance Planning (TAP), first introduced by Bloch and Salzman~\cite{bloch2024offlinetaskassistanceplanning},  we are given two robots called the task robot (\(\taskR\)) and the assistance robot (\(\assistR\)). 
The task robot needs to execute a task (e.g., inspecting an underground mine where there is limited communication) which requires moving in a known workspace.
The assistance robot can provide assistance (e.g., serve as a communication relay outside the mine in strategic locations such as shafts  where there is communication to \(\taskR\)) which is a function of the spatial configuration of both robots.\footnote{Importantly, assistance refers to support that does not influence task execution.} 
Because completing a task requires $\taskR$ to continuously move in the workspace,~$\assistR$ may also need to relocate its position in order to maximize assistance provided.
Bloch and Salzman~\cite{bloch2024offlinetaskassistanceplanning} consider the restricted setting where $\taskR$'s trajectory is fixed and we only need to plan the trajectory of $\assistR$.
In this work we consider a generalization of their problem (formally introduced in Sec.~\ref{sec:defs}) where we need to simultaneously compute the trajectories of~$\taskR$ and~$\assistR$ while maximizing the portion of \taskR’s path
for which assistance is provided. We call this \emph{Joint Task Assistance Planning} (\textsc{JointTAP}).

\figPlacemnet{
\begin{figure}
  \centering
  \subfloat[]{\includegraphics[width=0.6\columnwidth]{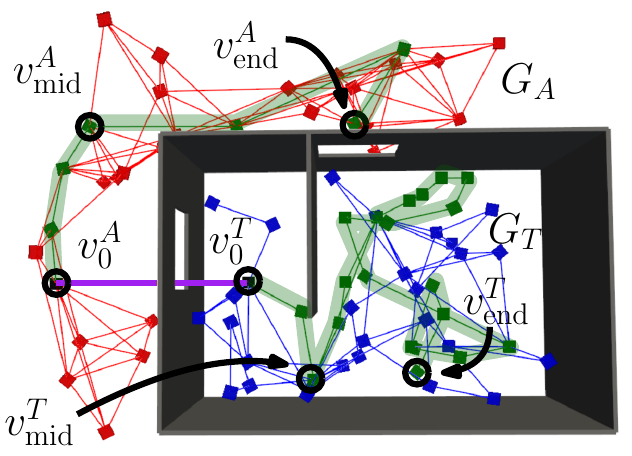}}\label{fig:a}
  \vspace{-2mm}
  \\  
  \subfloat[\OS{UPDATE}]{\includegraphics[width=0.3\columnwidth]{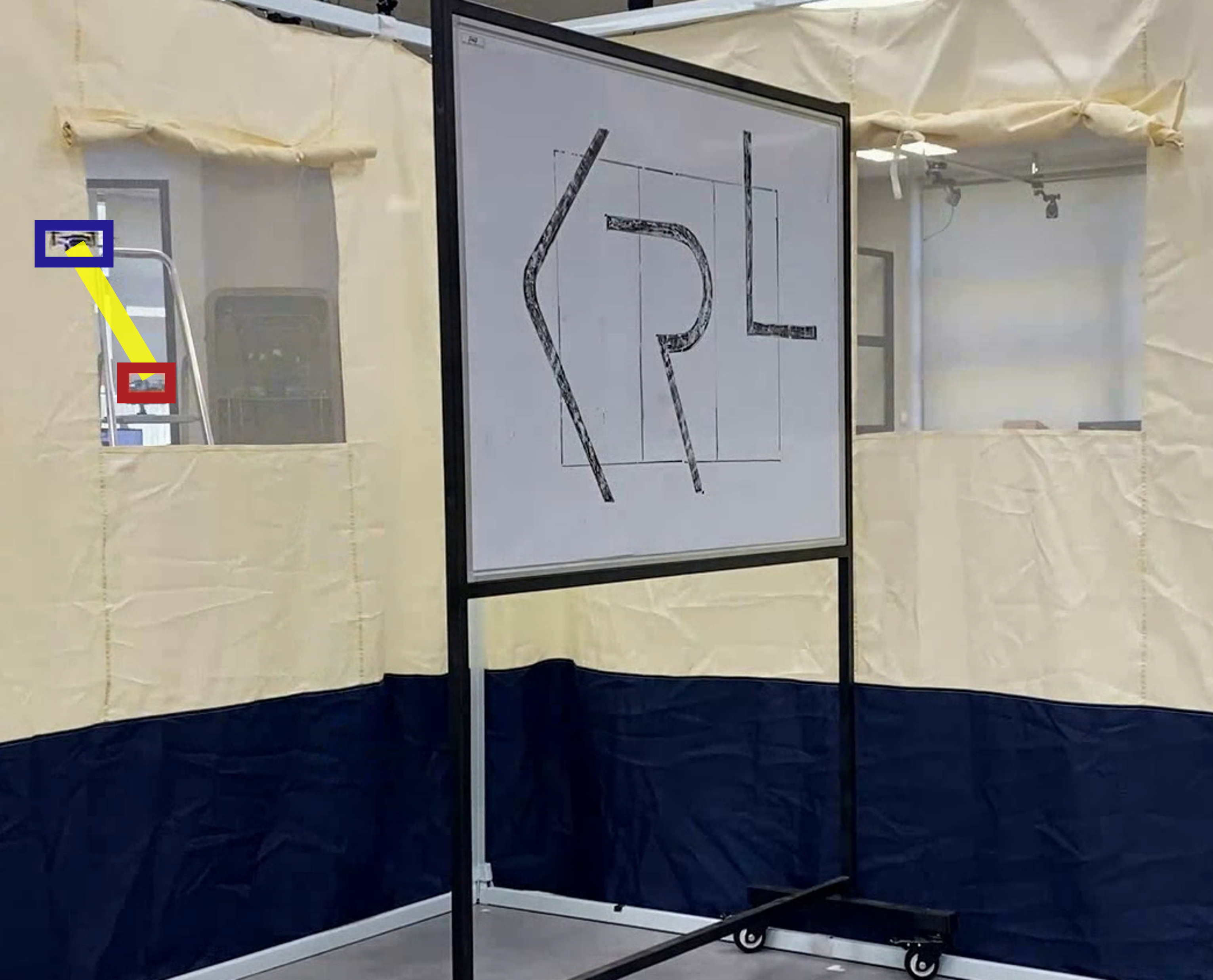}\label{fig:b}} \hfill
  \subfloat[]{\includegraphics[width=0.3\columnwidth]{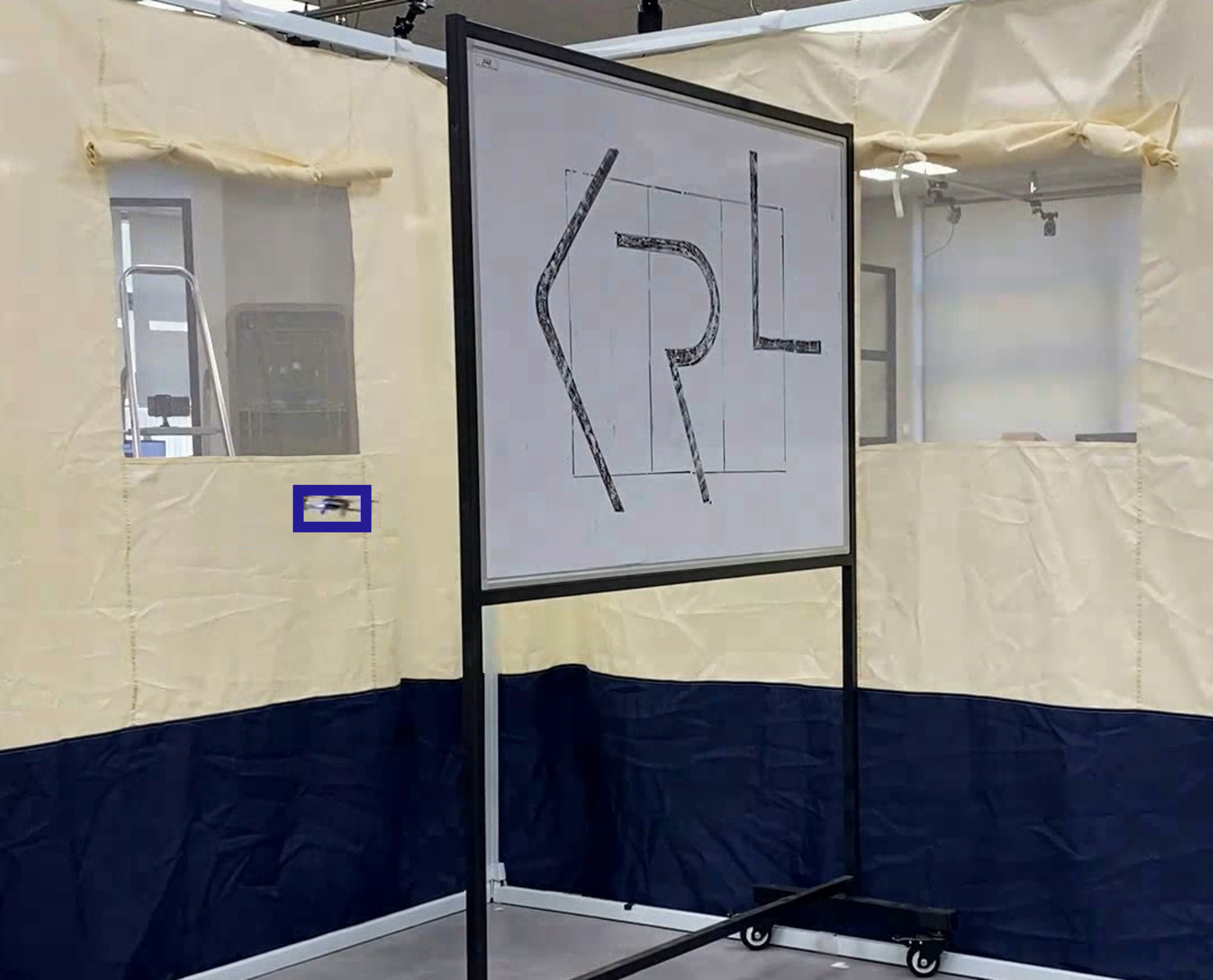}\label{fig:c}} \hfill
  \subfloat[]{\includegraphics[width=0.3\columnwidth]{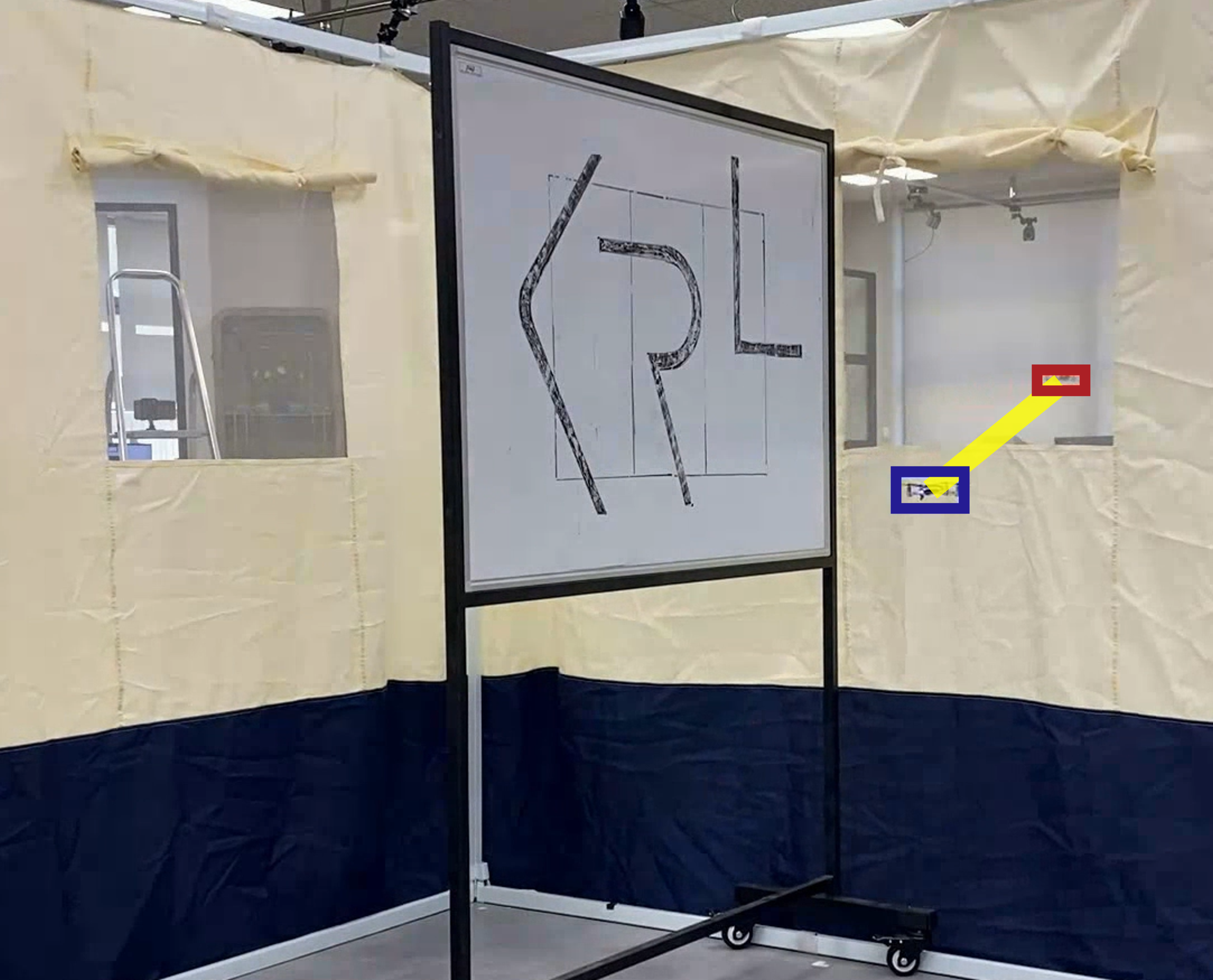}\label{fig:d}}
  \caption{(a) Illustration of motivating application where $\taskR$ (blue) and $\assistR$ (red) need to maximize accumulated line-of-sight (LOS) (purple) while moving on their respective roadmaps $G_{T}$, $G_{A}$. Computed paths depicted in green. 
  (b)-(d) Snapshots of path execution on Crazyflie drones for $t=0,0.3,1$ where
  $\taskR$ is located at $v_0^T, v_{0.3}^T, v_{\text{end}}^T$ and $v_0^A, v_{0.3}^A, v_{1}^A$, respectively.
  As $\taskR$ is required to move between two parts of the room, $\assistR$ needs to strategically reposition itself causing a temporary lack of LOS at t=0.3 (thus, $v_{0.3}^T$ is no visible in (b)).}
    \label{fig:motivating-example}
    \vspace{-3mm}
\end{figure}
}{
\begin{figure*}[t!]
  \centering
  \subfloat[]{\includegraphics[height=2.75cm]{Figs/room.pdf}\label{fig:motiva_1}}%
  \hfill
  \subfloat[]{\includegraphics[height=2.5cm]{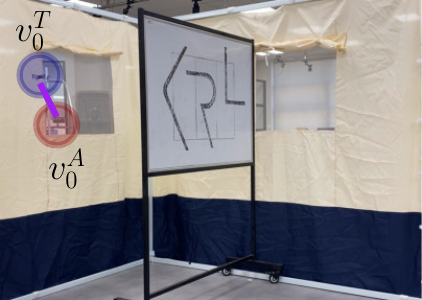}\label{fig:motiva_2}}%
  \hfill
  \subfloat[]{\includegraphics[height=2.5cm]{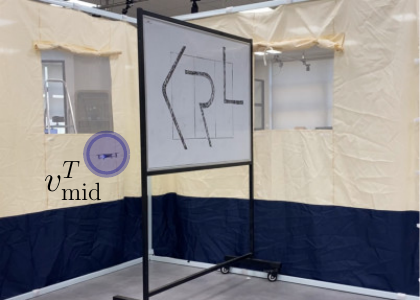}\label{fig:motiva_3}}%
  \hfill
  \subfloat[]{\includegraphics[height=2.5cm]{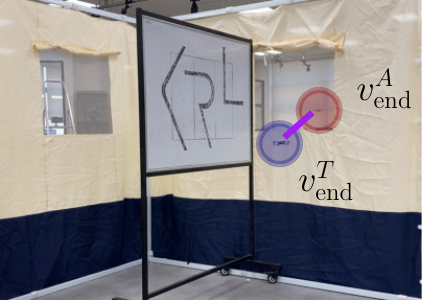}\label{fig:motiva_4}}%
  \caption{%
  (a) Illustration of motivating application where $\taskR$ (blue) and $\assistR$ (red) need to maximize accumulated line-of-sight (LOS) (purple) while moving on their respective roadmaps $G_{T}$, $G_{A}$. Computed paths depicted in green. 
  (b)-(d) Snapshots of path execution on Crazyflie drones for three timestamps $t_0=0, t_{\rm mid}$ and $t_{\rm end}=1$ where
  $\taskR$ is located at $v_0^T, v_{\rm mid}^T, v_{\rm end}^T$ and $v_0^A, v_{\rm mid}^A, v_{\rm end}^A$, respectively.
  As~$\taskR$ is required to move between two parts of the room, $\assistR$ needs to strategically reposition itself causing a temporary lack of LOS at $t_{\rm mid}$ (thus, $v_{\rm mid}^T$ is no visible in (b)).
  }
  \label{fig:motivating-example}
  \vspace{-5mm}
\end{figure*}
}
The \textsc{JointTAP} can be used to model a variety of problems where assistance (e.g., visual perception and or maintaining communication) plays a pivotal role.
One example can be found in settings where depth perception is critical, such as a tele-operated task in a confined environment, where an assistance robot equipped with a secondary camera can position itself at a different angle, helping a human operator judge distances more accurately~\cite{xiao2020tetheredaerialvisualassistance}.
As a second example, illustrated in Fig.~\ref{fig:motivating-example}, we consider search-and-rescue (SAR) operations where standard transmission methods may fail~\cite{lyu2023unmanned,murphy2008search}. Consequently, assistance can take the form of  a line-of-sight (LOS) communication link.
In such scenarios, an assistance robot positioned outside the environment can relay communication by maintaining LOS with the operating robot while  the task robot complete its mission. Thus, the mission-planner's objective is to jointly plan the path of both~$\taskR$ and $\assistR$ in order to (i) complete the task and (ii)~maximize assistance (i.e. maintaining communication).

Unfortunately, the TAP problem is already computationally hard~\cite{bloch2024offlinetaskassistanceplanning} rendering exact approaches to the \textsc{JointTAP} problem intractable. 
To this end, here we focus on the algorithmic foundations of the \textsc{JointTAP} problem and consider the offline discrete setting in which we compute the trajectories for both  $\taskR$ and $\assistR$ on predefined discrete motion-planning roadmaps before execution (similar to pre-operative planning in surgical setting---see, e.g.,~\cite{AlterovitzHK25}) and leave online \textsc{JointTAP} planning for future work.
%


Our contributions include the formulation of the \textsc{JointTAP} problem as well as an efficient Branch and Bound (BnB)-based algorithm that outperform baseline by up to two orders of magnitude. Key to the efficiency of our algorithm is a flow-based linear programming upper bound that allows to prune entire subregions of the search space. As candidate plans of the BnB are highly similar, we also introduce an incremental algorithm for optimizing subproblems and use it within our (BnB)-based algorithm. This additional optimization gives an extra speed up of up to {$3\times$}.
We demonstrate our algorithm in simulation on a 4-DOF planar manipulator and a {4-DOF} model of a drone in a 3D room as well as in the lab using {Crazyflie 2.1+ quadrotors} (Fig.~\ref{fig:motivating-example}).

\section{Related Work}
In this section, we consider broadly related work including visual assistance, inspection planning, and assisting agents in collaborative settings which bare resemblance to the TAP and \textsc{JointTap} problems.
We defer discussing the specific work of Bloch and Salzman~\cite{bloch2024offlinetaskassistanceplanning} until after the formal problem definition (Sec.~\ref{sec:defs}) and overview the algorithmic background they introduced (and we build upon) throughout the paper.

When the assistance takes the specific form of \emph{visual assistance}, our problem bares resemblance to problems related to robot target detection and tracking such as coverage, pursuit–evasion and surveillance~\cite{robin:hal-01183372}.  
Generally speaking, these can be split into collaborative and adversarial settings.
In the former, existing works either 
(i)~consider low-dimensional systems (see, e.g.,~\cite{LagunaB19,LaValleGBL97}), in contrast to the high-dimensional configuration spaces which motivate our work
or
(ii)~consider a relatively uncluttered environment and a fixed task trajectory as  in the case of planning camera motions (see, e.g.,~\cite{NieuwenhuisenO04a,rakita2018autonomous}).
In the latter,  adversarial setting (see, e.g.,~\cite{chung}),  one group of robots tracks another, whereas we focus on the cooperative setting in which the task and assistance robots work in concert.

Another topic that is closely related to visual assistance is inspection planning~\cite{FKSA19,FuSA21} or {coverage planning}~\cite{almadhoun2016survey,galceran2013RAS}.  
In these problems a robot (or team of robots~\cite{ropero2019terra}) are tasked with 
maximizing the coverage of a static area of interest. 
This differs from our setting wherein the temporal and sequential nature of assistance is paramount; it is not enough to simply see a region, but rather to assist the teammate in the workspace for as long as possible.

Finally, 
our work can be seen as an instance of problems studying  how agents can evaluate both their need for help and their ability to support others with the most closely related problem being computing the {Value of Assistance} (VOI)~\cite{VOI23,VOI23b}.
VOI calls for assessing (and computing) the gain obtained by  offering assistance  at a particular stage of the task-robot's execution.
In contrast to our setting, this approach limits assistance to a single point along the task-robot's trajectory, concentrating only on identifying the most beneficial location for intervention.

\tikzset{
  block/.style   = {rectangle, minimum width=3cm, minimum height=1cm,
                    draw, fill=gray!10, text centered},
  decision/.style= {diamond,  aspect=2, inner sep=1pt,
                    draw, fill=gray!15, text centered},
  >={Stealth},
  line/.style    = {->, thick},
}

\section{Problem Definitions}
\label{sec:defs}
Recall that in our setting we are given two robots $\taskR$ and $\assistR$ called the \emph{task robot} and \emph{assistance robot}, respectively.
Their possible motions are captured via graphs $G_T = (V_T, E_T)$ and $G_A = (V_A, E_A)$ called the \emph{task graph} and \emph{assistance graph}, respectively. These represent roadmaps embedded in the robot's configuration spaces~\cite{Salzman19} where vertices and edges of $G_T$ and $G_A$  correspond to configurations and transitions of robots. 
Furthermore, 
we are given start and goal vertices $v_0^T, v_{\text goal}^T \in V^T$ for $\taskR$
as well as
a start vertex \(v_0^A\in V_A\) for $\assistR$ which encode the set of valid paths in $G_T$ and $G_A$, respectively.
Each edge~$e \in E_A \cup E_T$ has a length $\ell(e)$.
We assume that
moving along an edge $e$ takes time  that's identical to $\ell(e)$
and 
that time and edge length are normalized to~$[0,1]$.

We model  \assistR's ability to provide assistance to  \taskR via a \emph{assistance function} 
$\A: V_A \times V_T \rightarrow \{0,1\}$
such that 
if~\assistR is located at $u_A$ and~\taskR is located at~$u_T$
then 
$\A(u_A,u_T) = 1$ and $\A(u_A,u_T) = 0$ indicates that~\assistR can and can't provide assistance to \taskR, respectively.
To model assistance when~\assistR moves along edges, we extend~$\A$ as follows: 
Given edge $e = (u,v) \in E_A$ traversed by~\assistR, we \emph{associate}~\assistR with 
$u$ and $v$ if it is located along the first and second half of $e$, respectively. The notion of robot association is defined analogously for~\taskR.
Now, when \assistR and \taskR traverse edges $e_A$ and $e_T$, respectively, the assistance is defined with respect to their associated vertices.

{
As assistance at edges is a function of assistance at vertices, it will be useful to 
extend the length function~\(\ell\):
Given path~\(\pi=\langle v_0,\ldots,v_n\rangle\) in \(G\in \{G_T,G_A\}\) we define~$\ell_\pi(v_i,v_j)$ to be the length of a path from~$v_i$ to~$v_j$.
That is, $\ell_\pi(v_i,v_j):= \sum_{h=i}^{j-1}{\ell(v_h, v_{h+1})}$.
To simplify the definition we set \(\ell_\pi(v_{-1},v_0):=0\), \(\ell_\pi(v_{k},v_{k+1}):=0\), and omit $\pi$ when understood from context. 
%
%
Moreover, we denote~\(\delta'(v, v')\) to be the shortest path length excluding the first and last half-edges between $v$ and $v'$ among \emph{all} paths in $G$.
}

Recall that 
$\taskR$ and $\assistR$ move at constant speed and that time $t$ is normalized such that $t \in [0,1]$.
However, we also assume that they can stop at graph vertices. Thus, to determine their location along a given path, we need to specify how long they stop at each vertex along the path. This is captured using the notion of a \emph{timing profile}.\footnote{The notion of a timing profile is relevant both to paths of $\taskR$ in $G_T$ and to paths of \assistR in $G_A$.}
%
{
\begin{dfn}[Timing Profile]
    Let $\pi = \langle v_0,\ldots, v_k\rangle$ be a path in $G \in \{G_T, G_A\}$. 
    A \emph{timing profile} of $\pi$ is a sequence of timestamps~$\T = \langle t_0, \dots, t_{k-1} \rangle $ s.t.
    $t_{i+1} \geq t_i + 0.5\cdot \ell(v_i, v_{i+2})$
    for all \(0\le i < k\), where we assume \(t_{-1} = 0, t_k = 1\).
\end{dfn}
}

\noindent
\textbf{Assumption.}
As the ultimate goal of \taskR is to complete its task, we assume that its timing profile is fully determined for a given path~$\pi_T$ such that it is not delayed unnecessarily at vertices. Thus, for a given path we denote the corresponding  timing profile by~$\T_T(\pi_T)$. 
This also means that once a path is fixed, we can associate each assistance vertex~\(v_A \in V_A\) with a set of \emph{time intervals}~\(\I^\A_{\pi_T}(v_A)\), called assistance intervals, capturing the time windows during which \(\taskR\) is present at a vertex \(v_T\) assisted by \(v_A\). This assumption can easily be relaxed but we avoid this to simplify the exposition.

Recall that $\mathcal{A}$ indicates whether assistance can or can't be provided given two vertices. As we will see, it will be convenient to extend~$\A$ to time intervals. Specifically, and with a slight abuse of notation, we set
$\A: V_A \times V_T \times 2^{[0,1]} \times 2^{[0,1]} \rightarrow [0,1]$
such that
$\A(u_A, u_T, I_A, I_T):=\A(u_A, u_T) \cdot \left|I_A \cap I_T\right|$. 
This can be interpreted as the amount of assistance provided when 
\assistR resides in locations associated with $u_A$ in the time interval $I_A$ 
while 
\taskR resides in locations associated with $u_T$ in the time interval $I_T$.

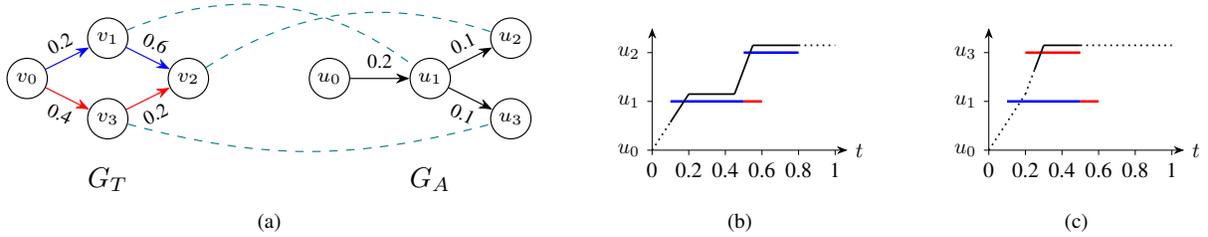
\begin{figure*}[t]
\centering

\subfloat[\label{fig:task-assist-graphs}]{
\resizebox{0.4\textwidth}{!}{%
\begin{tikzpicture}[>=Stealth,
    node/.style={circle,draw,minimum size=5mm,inner sep=0pt,font=\scriptsize}]

  \node[node] (v0) at (0, 0)  {$v_0$};
  \node[node] (v1) at (1,0.5)  {$v_1$};
  \node[node] (v3) at (1,-0.5) {$v_3$};
  \node[node] (v2) at (2,0)  {$v_2$};

\draw[->, blue] (v0) -- node[midway, sloped, above, black] {\scriptsize 0.2} (v1);
\draw[->, red] (v0) -- node[midway, sloped, below, black] {\scriptsize 0.4} (v3);
\draw[->, blue] (v1) -- node[midway, sloped, above, black] {\scriptsize 0.6} (v2);
\draw[->, red] (v3) -- node[midway, sloped, below, black] {\scriptsize 0.2} (v2);

  \node at (1,-1.25) {$G_T$};

  \node[node] (u0) at (3.75, 0)  {$u_0$};
  \node[node] (u1) at (5, 0)     {$u_1$};
  \node[node] (u2) at (6, 0.5)     {$u_2$};
  \node[node] (u3) at (6,-0.5)     {$u_3$};

  \draw[->] (u0) -- node[midway, above] {\scriptsize 0.2} (u1);
  \draw[->] (u1) -- node[midway, sloped, above] {\scriptsize 0.1} (u2);
  \draw[->] (u1) -- node[midway, sloped, below] {\scriptsize 0.1} (u3);

  \node at (5,-1.25) {$G_A$};

\draw[teal, dashed, bend right=15] (v3) to (u3);
\draw[teal, dashed, bend left=30]  (v1) to (u1);
\draw[teal, dashed, bend left=25]  (v2) to (u2);
\end{tikzpicture}
}%
}
\hfil
\subfloat[\label{fig:timing-up}]{
\raisebox{.1\height}{%
\resizebox{0.2\textwidth}{!}{%
\begin{tikzpicture}[scale=0.8]
  \begin{scope}
    \draw[->] (0,0) -- (4,0);
    \draw[->] (0,0) -- (0,2.5);
    \node at (4.25,0) {$t$};

    \foreach \x/\xtext in {0/0, 0.75/0.2, 1.5/0.4, 2.25/0.6, 3/0.8, 3.75/1} {
      \draw (\x,0) -- (\x,-0.1) node[below] {\xtext};
    }

    \foreach \y/\ylabel in {0/u_0, 1/u_1, 2/u_2} {
      \draw (0,\y) -- (-0.1,\y) node[left] {$\ylabel$};
    }

    \draw[blue, very thick] (0.375,1) -- (1.875,1);
    \draw[blue, very thick] (1.875,2) -- (3,2);
    \draw[red,  very thick] (1.875,1) -- (2.25,1);

    \draw[black, thick]           (0.375,0.575) -- (0.75,1.15);
    \draw[dotted, black, thick]   (0,0) -- (0.375,0.575);
    \draw[black, thick]           (0.75,1.15) -- (1.6875,1.15);
    \draw[black, thick]           (2.0625,2.15) -- (3,2.15);
    \draw[dotted, black, thick]   (3,2.15) -- (3.75,2.15);
    \draw[black, thick]           (1.6875,1.15) -- (1.875,1.65);
    \draw[black, thick]           (1.875,1.65) -- (2.0625,2.15);
  \end{scope}
\end{tikzpicture}
}%
}%
}
\hfil
\subfloat[\label{fig:timing-down}]{
\raisebox{.1\height}{%
\resizebox{0.2\textwidth}{!}{%
\begin{tikzpicture}[scale=0.8]
  \begin{scope}
    \draw[->] (0,0) -- (4,0);
    \draw[->] (0,0) -- (0,2.5);
    \node at (4.25,0) {$t$};

    \foreach \x/\xtext in {0/0, 0.75/0.2, 1.5/0.4, 2.25/0.6, 3/0.8, 3.75/1} {
      \draw (\x,0) -- (\x,-0.1) node[below] {\xtext};
    }

    \foreach \y/\ylabel in {0/u_0, 1/u_1, 2/u_3} {
      \draw (0,\y) -- (-0.1,\y) node[left] {$\ylabel$};
    }

    \draw[blue, very thick] (0.375,1) -- (1.875,1);
    \draw[red,  very thick] (1.875,1) -- (2.25,1);
    \draw[red,  very thick] (0.75,2)  -- (1.875,2);

    \draw[black, thick]           (1.125,2.15) -- (1.875,2.15);
    \draw[dotted, black, thick]   (1.875,2.15) -- (3.75,2.15);
    \draw[dotted, black, thick]   (0,0) -- (0.75,1.15);
    \draw[black, thick]           (0.9375,1.65) -- (1.125,2.15);
    \draw[dotted, black, thick]   (0.75,1.15) -- (0.9375,1.65);
  \end{scope}
\end{tikzpicture}
}%
}%
}

\caption{Illustrative example and induced timing. 
(a) Task graph \(G_T\) with start and goal vertices $v_0,v_2$ and assistance graph~\(G_A\) with start vertex $u_0$ and assistance function~\(\mathcal{A}\) (teal dashed). 
(b) Intervals from \(\pi_T^\uparrow=\langle v_0,v_1,v_2\rangle\) (blue) together with a timing profile on \(\pi_A^\uparrow\) (black). 
(c) Intervals from \(\pi_T^\downarrow=\langle v_0,v_3,v_2\rangle\) (red) with a timing profile on \(\pi_A^\downarrow\). 
Solid black segments indicate times where assistance can be performed and dotted segments where it cannot.}
\label{fig:full-width-three-subfigs}
\vspace{-5mm}
\end{figure*}

Using this extended definition of $\A$, we define the notion of \emph{reward} which measures the amount of assistance~\assistR provides to \taskR while each is traversing a given path.
\begin{dfn}[Reward]
    Let $\pi_A = \langle v_0^A,\ldots,v_k^A\rangle$ be a path in \(G_A\), 
    $\pi_T = \langle v_0^T,\ldots,v_m^T\rangle$ be a path in \(G_T\),
    $\T_A = \langle t_0^A,\ldots,t_{k-1}^A\rangle$ be the timing profile of~$\pi_A$
    and
    let $\T_T = \langle t_0^T,\ldots,t_{m-1}^T\rangle$ be the timing profile of $\pi_T$.
    The \emph{reward} of paths~$\pi_A, \pi_T$ using timing profiles $\T_A, \T_T$
    is defined as
    \begin{equation}
        \label{eq:reward}
                \R_\A(\pi_A, \pi_T, \T_A, \T_T):=\sum_{i=0}^{k}\sum_{j=0}^{m}{\A(v_i^A, v_j^T, \T_A^i, \T_T^j)}.
    \end{equation}
    Here, let $\T^i_A = [t_{i-1}^A,t_i^A]$, and define $\T^i_T$ analogously.

\end{dfn}


We are finally ready to formally define our problems of interest. 
We start with the restricted timing-focused problem and continue to our general problem. 

\begin{prob}[OTP]
\label{prob:OTP}
    Let $\pi_A,\pi_T$ be paths in $G_A,G_T$, respectively.
    Let $\mathcal{T}(\pi_A)$ be the set of all possible timing profiles over $\pi_A$
    and 
    $\T_T$ be the timing profile of~$\pi_T$.
    The \emph{Optimal Timing Profile (OTP)} problem calls for computing a timing profile~$\T^*_A$ for $\pi_A$ whose reward is maximal.
    Namely, compute $\T^*_A$ such that
    $$
        \T^*_A \in \argmax_{\T_A \in \mathcal{T}(\pi_A)} \R_\A(\pi_A, \pi_T, \T_A, \T_T(\pi_T)).
    $$
\end{prob}

\begin{prob}[\textsc{JointTAP}]
\label{prob:JointTAP}
    Let $v^T_0 \in V_T$ and $v^A_0 \in V_A$ be start vertices for \taskR and \assistR, respectively, 
    let $\Pi_T(v_0^T, v_{\text{end}}^T)$ 
    be the set of paths in~$G_T$ 
    starting from $v_0^T$ and ending at~$v_{\text{end}}^T$
    and 
    let $\Pi_A(v_0^A)$ 
    be the set of paths in~$G_A$
    starting from $v_0^A$.
    Furthermore, let $\mathcal{T}(\pi_A)$ be the set of all possible timing profiles over path~$\pi_A$ 
    and    
    recall that $\T_T(\pi_T)$ is the induced timing profile of path~$\pi_T$ in~$G_T$.
    The \emph{Joint Task Assistance Planning (\textsc{JointTAP})} problem calls for computing a path~$\pi^*_T$ in \(G_T\), path $\pi^*_A$ in \(G_A\) and a timing profile~$\T^*_A$  whose reward is maximal.
    Namely, compute $\pi^*_T,\pi^*_A, \T^*_A$ s.t.
    $$
    \pi_T^*,\pi^*_A,\T^*_A \in 
        \argmax_{\substack{
            \pi_A \in \Pi_A ( v_0^A), \\ 
            \pi_T \in \Pi_T(v_0^T, v_{\text{end}}^T), \\ 
            \T_A \in \mathcal{T} \left(\pi_A \right)}} 
        \R_\A \left(\pi_A, \pi_T, \T_A, \T_T(\pi_T) \right).
    $$

\end{prob}



Finally, we call the special case of \textsc{JointTAP} (Prob.~\ref{prob:JointTAP}), where the task path~$\pi_T$ is fixed the \emph{Assistance Optimal Timing Profile (\textsc{AssistanceOTP})} problem.\footnote{The \textsc{AssistanceOTP} was studied in~\cite{bloch2024offlinetaskassistanceplanning} under the name  \textsc{OPTP}. }

\begin{example}
\label{ex:notation}
To illustrate \textsc{JointTAP}, consider the instance shown in Fig.~\ref{fig:task-assist-graphs}.
On the task graph \(G_T\), we consider two candidate paths: \(\pi_T^\uparrow = \langle v_0, v_1, v_2 \rangle\) and \(\pi_T^\downarrow = \langle v_0, v_3, v_2 \rangle\), which induce the timing profiles \(\T_T(\pi_T^\uparrow) = \langle 0, 0.1, 0.5, 0.8 \rangle\) and \(\T_T(\pi_T^\downarrow) = \langle 0, 0.2, 0.5, 0.6 \rangle\), respectively. These profiles, together with the assistance relation~\(\A\), determine the corresponding intervals on~\(G_A\) are depicted in Fig.~\ref{fig:full-width-three-subfigs}. 

Path \(\pi_T^\uparrow\) and assistance $\A$ induce assistance intervals \([0.1, 0.5]\) on \(u_1\) and \([0.5, 0.8]\) on \(u_2\), depicted in blue in Fig.~\ref{fig:timing-up},\ref{fig:timing-down}. Now, $\assistR$ can follow path \(\pi_A^\uparrow\) with timing profile \(\T_A^\uparrow = \langle 0, 0, 0.45, 1 \rangle\) (Fig.~\ref{fig:timing-up}), covering both intervals and yielding a total reward of \(\R_\A(\pi_A^\uparrow, \pi_T^\uparrow, \T_A^\uparrow, \T_T(\pi_T^\uparrow)) = 0.7\). 

Alternatively, choosing \(\pi_T^\downarrow\) yields shorter intervals and results in lower reward of \(0.25\).
\end{example}

\section{Algorithmic Approach}
Recall that to solve the \textsc{JointTAP} problem (Prob.~\ref{prob:JointTAP}), we need to compute 
a task path~$\pi_T$, 
an assistance path~$\pi_A$, 
and a timing profile for~$\pi_A$ which jointly maximize the total reward.
Bloch and Salzman~\cite{bloch2024offlinetaskassistanceplanning} introduced an efficient solver for \textsc{AssistanceOTP} which, given a 
task path~$\pi_T$,  computes 
an assistance path~$\pi_A$, 
and a timing profile for~$\pi_A$ which jointly to maximize total reward.
Thus, a straw man approach, 
could be enumerating  all task paths and, for each one solve the corresponding \textsc{AssistanceOTP} problem. 
As we will see in Sec.~\ref{sec:experiments},  while solving the problem optimally, this approach is impractical due to its exponential runtime.

The key shortcomings of this approach are that 
(i)~task paths are naively enumerated
and that 
(ii)~the corresponding \textsc{AssistanceOTP} problem is recomputed from scratch.

To systematically and efficiently enumerate the set of task paths, we suggest to employ a nested branch-and-bound (BnB) framework~\cite{brown2015globally,campbell2018globally,yang2015go}.
BnB is a general algorithmic framework for solving combinatorial-optimization problems~\cite{clausen1999branch}. 
It explores the solution space by dividing it into smaller subproblems (branching) and uses bounds to prune regions that cannot contain better solutions (bounding). This often allows finding on optimal solution efficiently without iterating over every possible solution.
Conceptually, nested BnB is a hierarchical optimization framework that applies BnB recursively at outer and inner levels of an optimization problem. 
The outer BnB explores a high-level search space (in our setting, this well be selecting a task path~\(\pi_T\) in~\(G_T\)), while the inner BnB solves a dependent subproblem (in our setting, this will be  computing the optimal assistance path~\(\pi_A\) and timing profile for each outer solution~\(\pi_T\)). This structure enables efficient pruning based on upper bounds at both levels, allowing the algorithm to avoid exhaustive search while preserving optimality.
We detail this step in Sec.~\ref{sec:nested-bnb}.

A key optimization we employ within our BnB instantiation builds upon the observation that we solve multiple, highly similar \textsc{OTP} problems (Prob.~\ref{prob:OTP}). To avoid recomputing solutions from scratch, we suggest in Sec.~\ref{sec:inc-otp} an incremental approach for solving an \textsc{OTP} problem~$P$ given that we have a solution to a similar Problem~$P'$ .



\section{Incrementally Solving the \textsc{OTP} Problem}
\label{sec:inc-otp}

\ignore{As we will see, our approach for solving the \textsc{JointTAP} problem will require solving the \textsc{OTP} problem for a sequence of highly-similar queries that only differ with respect to a small number of assistance intervals. 
Clearly, one could solve this sequence of similar \textsc{OTP} problems from scratch one at a time but this is wasteful from a computational point of view.
To this end, we suggest an approach for solving an \textsc{OTP} problem $P$ given that we have a solution to a similar Prob.~$P'$.
The rest of this section formalizes this idea.}

Before we describe our approach, and to simplify this section, we present an alternative formulation to the \textsc{OTP} problem (Prob.~\ref{prob:OTP}):
First, as both paths $\pi_T,\pi_A$ and the timing profile~$\T_T$ of~\taskR are fixed, 
we can compute for each vertex $v \in \pi_A$  the set of time intervals during which \assistR can assist \taskR and  denote this set as $\I^\A_{\pi_T}(v)$.
In addition, we use $\mathcal{V}(I)$ to denote the vertex associated with interval $I$ (i.e., $\forall I \in \I^\A_{\pi_T}(v),~\mathcal{V}(I) = v$).

In this specific case, the reward can be written as: 
\vspace{-1.5mm}
\begin{equation}
    \label{eq:reward2}
        \R_\A(\pi_A, \pi_T, \T_A, \T_T):=
            \sum_{i=0}^{k}\sum_{I \in \I^\A_{\pi_T}(v_i) }
                \vert
                    \T_A^i \cap I
                \vert.
\end{equation}
Importantly, the original definition (Eq.~\eqref{eq:reward}) is equivalent to this new one (Eq.~\eqref{eq:reward2}).
In the remainder of this section we will use this reward definition  and assume that an \textsc{OTP} problem is given in the form of 
an assistance path $\pi_A$ and  the intervals~$\I^\A_{\pi_T}(v)$ for each vertex $v \in \pi_A$.\footnote{This definition is the one used by Bloch and Salzman~\cite{bloch2024offlinetaskassistanceplanning} since they considered the setting where $\pi_T$ is fixed thus avoiding the need to reason about timing profiles of two paths. Our ultimate goal, solving the \textsc{JointTAP} problem requires reasoning about both paths which is why  the reward defined (Eq.~\ref{eq:reward}) is in a general form.}

\subsection{Algorithmic background}
Our approach relies on the notion of \emph{critical times} and \emph{time-reward pairs}~\cite{bloch2024offlinetaskassistanceplanning}.
Conceptually, the set of critical times capture two key transitions between intervals: arriving exactly at the start of an assistance interval or leaving exactly at its end.\footnote{Here, the phrase ``$\assistR$ arrives at vertex~$v$'' is used in terms of assistance (i.e., when~$\assistR$ can start to provide the assistance associated with vertex~$v$). Specifically, following our assistance model (Sec.~\ref{sec:defs}), 
when~$\assistR$ arrives at time $t$ to vertex \(v_i\) of  path \(\pi_A = \langle v_0, \ldots, v_n \rangle\), it actually starts providing assistance at time~\(t - \frac{1}{2}\ell(v_{i-1}, v_i)\).}
Now, given an \textsc{OTP} problem, one can show that it is sufficient to restrict the search for an optimal timing profile to combinations of these critical times, thereby discretizing the otherwise continuous problem. 

Critical times can then be used to solve the \textsc{OTP} problem as follows:
(i)~compute $\textsc{Ct}_i$, the set of all critical times associated with vertex $v_i$ of assistance path $\pi_A$
and 
(ii)~for each vertex~$v_i$ and  each critical time $t \in \textsc{Ct}_i$, compute best reward $r$ that is achievable by exiting~$v_i$ at time $t$. 
This can be computed efficiently by a dynamic-programming approach that maintains so-called \emph{time-reward pairs} associated with path vertices.
Such a pair $(t,r)$ associated with vertex $v_i \in \pi_A$ represents that $r$ is the  best reward achievable by exiting~$v_i$ at critical time $t \in \textsc{Ct}_i$.


Importantly, the critical times $\textsc{Ct}_i$ (which are precomputed before computing all time-reward pairs) require accounting for the start and end times of each interval in $v_i$ and of future vertices in $\pi_A$.
Consequently, a change in the interval set of any vertex may require updating the critical times at earlier vertices effectively necessitating a full rerun of the algorithm.

As we will see shortly, our incremental approach does not require precomputing all critical times which allows for efficient solution updates when a new interval is added.

\subsection{Algorithmic framework}

\subsubsection*{High-level approach}

To allow for incremental changes, our algorithm will maintain a data structure which we call the \emph{history}~$\His$ 
which is an ordered list of intervals $I_1, I_2\ldots $
ordered first by the order of vertices in~$\pi_A$ and then according to the interval's start time.
I.e., given two intervals $I, I'$  with start times $t$ and~$t'$, respectively such that $I$ appears before~$I'$ then 
either $\mathcal{V}(I)$ is before $\mathcal{V}(I')$ in~$\pi_A$
or alternatively, $\mathcal{V}(I) = \mathcal{V}(I')$ and thus  $t \leq t'$. 

%
%

%
Unlike the original algorithm, which precomputes the full set of critical times~\(\textsc{Ct}_i\) for each vertex, our approach computes critical times on demand. Instead of maintaining a global list of all critical times per vertex, each interval~$ \in \His$ stores only the relevant time-reward pairs in a set called~\textsc{Arrivals} that are associated with $I$. Importantly, $\textsc{Arrivals}(I)$ contains time-reward pairs whose critical times are from intervals before $I$ in $\His$. The interval-centric approach (in contrast to the vertex-centric approach of Bloch and Salzman) localizes computation as each interval maintains only the critical times pertinent to itself.

When a new interval~\(I_\text{new}\) is added to~\(\His\), we compute $\textsc{Arrivals}(I_{\text{new}})$
based on preceding intervals that can reach it. This is done by considering the critical times: either by arriving to~\(I_\text{new}\) at its start time or leaving a preceding interval at its end time. After this update, we similarly update every interval in~\(\His\) that comes after \(I_\text{new}\), to account for new possible transitions originating from  \(I_\text{new}\).

{Once~\(\His\) is updated, each interval $I \in \His$ stores the set~\(\textsc{Arrivals}(I)\) containing the possible arrival times and their corresponding rewards. This allows us to extract, for each interval, the optimal reward achievable up to that point.
%


\subsubsection*{Algorithmic details}

\begin{algorithm}[t]
\caption{Incremental OTP Procedures}
\label{alg:incremental-otp}
\begin{algorithmic}[1]
\Function{IncOTP}{$\His,  \I, \pi_A$}
\For{$I \in \I$} \label{line:forhist}
\State \(\His \gets\texttt{InsertInterval}({\His,I})\)  \label{line:callinsert}
\EndFor
\State \Return \(\texttt{FindMaxReward}({\His}),\His\) \label{line:callmaxreward}
\EndFunction
\vspace{0.5em}

\Function{InsertInterval}{$\His,  I_\text{new}$}

\State $\His.\texttt{Insert}\left(I_\text{new}\right)$  \label{line:incotp-insert} \label{line:incotp-init1}
\State \(\His_\text{prev},\His_\text{post}\gets \texttt{SplitHistory}(\His, I_\text{new})\) \label{line:split1}
\State \(\His' \gets \His_\text{prev}\) \label{line:incotp-init2}
\For{$I_\text{dest}=[\alpha_\text{dest}, \beta_\text{dest}]\in\His_\text{post}$} \label{line:incotp-fordest} 
    \State \(\His_\text{prev}',\His_\text{post}'\gets \texttt{SplitHistory}(\His', I_\text{dest})\)
    \For{$I_\text{src}=[\alpha_\text{src}, \beta_\text{src}]\in\His_\text{prev}'$} \label{line:incotp-forsrc} 
        \State \(\texttt{AddReward}({I_\text{src},I_\text{dest},\alpha_\text{dest}})\) \label{line:incotp-arrivestart} 
        \State \(\texttt{AddReward}({I_\text{src},I_\text{dest},\beta_\text{src}+\ell (\mathcal{V}(I_\text{src}),\mathcal{V}(I_\text{dest}))})\) \label{line:incotp-leaveend}  
    \EndFor
    \State $\His'.\texttt{Insert}(I_\text{dest})$ \label{line:incotp-appendintervals}
\EndFor
\State \Return $\His'$
\EndFunction

\end{algorithmic}
\end{algorithm}

We are finally ready to detail our algorithm outlined in Alg.~\ref{alg:incremental-otp}.
The main function \texttt{IncOTP} receives as input the history\footnote{Any history should include the interval \([0,0]\) at vertex \(0\) with initial time-reward pair of \((0,0)\).}~$\His$, a set of intervals $\I$ to add associated with assistance path $\pi_A$. The function incrementally adds\footnote{For simplicity, we only consider adding intervals but the entire approach can be used to remove intervals from $\His$. This is done by replacing the \texttt{insert} operation (Alg.~\ref{alg:incremental-otp}, Line~\ref{line:incotp-insert}) to \texttt{remove}.} the intervals from $\I$ to $\His$ (Lines~\ref{line:forhist}-\ref{line:callinsert} and detailed in  \texttt{insertInterval}) and then computes the reward associated with the best timing profile (Line~\ref{line:callmaxreward} and detailed in  \texttt{findMaxReward}).


Recall, that when adding interval $I_\text{new}$ into $\His$ (function \texttt{insertInterval}), we are only required to update the time-reward pairs for $I_\text{new}$ as well as for subsequent intervals in $\His$. 
Thus, after adding $I_\text{new}$ to $\His$ (Line~\ref{line:incotp-init1}), we split $\His$ into two sets:
$\His_\text{prev}, \His_\text{post}$ which contain all intervals that appear before and after~$I_\text{new}$ in~\(\His\), respectively (importantly,~$I_\text{new} \in \His_\text{post}$). This is implemented via the function \texttt{SplitHistory} (Line~\ref{line:split1}).

We then continue to iterate over all intervals in $\His_\text{post}$ in order to update their time reward pairs (Line~\ref{line:incotp-fordest}).
For each such interval $I_{\text{dest}} \in \His_\text{post}$ we consider all intervals  $I_{\text{src}} \in \His_\text{post}$ that lie before $I_{\text{dest}}$ (Line~\ref{line:incotp-forsrc}).
We then add time reward pairs to $\textsc{Arrivals}(I_{\text{dest}})$ corresponding to the critical times 
where
(i)~we arrive exactly at the start of \(I_{\text{dest}}\) (Line~\ref{line:incotp-arrivestart}), 
or where 
(ii)~we leave exactly at the end of \(I_{\text{src}}\) (Line~\ref{line:incotp-leaveend}).

In either case, updating $\textsc{Arrivals}(I_{\text{dest}})$ is done by the function \texttt{AddReward}(\(I_\text{src},I_\text{dest},t_\text{arrive}\)). 
{This function computes the reward of the optimal transition from \(I_\text{src}\) to \(I_\text{dest}\), assuming \(\assistR\) arrives to \(I_\text{dest}\) at \(t_\text{arrive}\).
It does so by computing the maximum reward achievable up to $t_\text{arrive}$ 
from the time-reward pairs in $\textsc{Arrivals}(I_{\text{src}})$. 
The resulting reward~$r$ is then paired with $t_\text{arrive}$ and added to 
$\textsc{Arrivals}(I_{\text{dest}})$. 
}


\subsubsection*{Complexity}
Let \(n\) denote the total number of intervals and~\(m\) the number of intervals that follow \(I_\text{new}\) in \(\His\). For each such interval \(I_\text{dest}\) (there are \(m\) of them), we iterate over all preceding intervals \(I_\text{src}\) in \(\His\), of which there are~\(\mathcal{O}(n)\). For each pair \((I_\text{src}, I_\text{dest})\), we invoke \texttt{AddReward} twice. 
\texttt{AddReward} computes the reward obtainable between these two intervals and performs insertion and lookup in the \(\textsc{Arrivals}\) list. Thus, each call to \texttt{AddReward} takes~$\mathcal{O}(n)$ time.
Consequently, the overall complexity for inserting an interval is \(\mathcal{O}(mn^2)\). }
This can be improved to  \(\mathcal{O}(mn\log n)\)
 using tailored  data structure and caching. Details omitted.


\subsubsection*{Correctness (sketch)}
To prove that Alg.~\ref{alg:incremental-otp} computes an optimal reward, note that for each interval \(I_i\) in history~\(\His = \langle I_0, \ldots, I_n \rangle\), the algorithm maintains a set of time-reward pairs \(\textsc{Arrivals}(I_i)\), where each pair corresponds to one of the two critical transition types from a preceding interval: arriving exactly at the start of \(I_i\) or leaving exactly at the end of the preceding interval.

Next, consider an optimal timing profile $\T_{\rm opt}$ that visits the sequence of intervals \(\langle I_{i_0}, \ldots, I_{i_k} \rangle\). Following Bloch and Salzman~\cite{bloch2024offlinetaskassistanceplanning}, we can assume that $\T_{\rm opt}$ only contains critical times.
That is, each transition between two consecutive intervals in the sequence either arrives at the start of the latter or leaves at the end of the former.

Because the algorithm populates each \(\textsc{Arrivals}(I)\) with both types of critical transitions from all former intervals, it must eventually include the correct time-reward pair \((t, r)\) for the final interval \(I_{i_d}\). This pair is obtained through a valid chain of transitions starting from \(\textsc{Arrivals}(I_{i_0})\), propagating forward through the sequence using only critical transitions. Since the timing and reward exactly match those guaranteed by Bloch and Salzman’s construction, we have the same sequence of intervals and the same set of critical transitions. Thus, the algorithm correctly recovers the optimal timing profile's reward.

\section{Nested Branch-and-Bound for \textsc{JointTAP}}
\label{sec:nested-bnb}

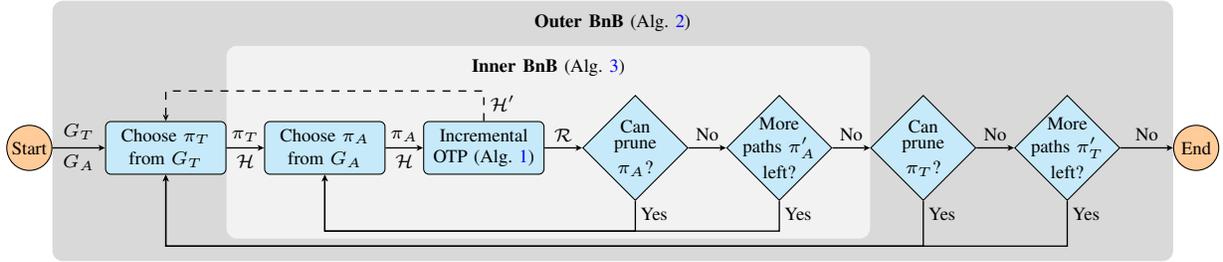
\begin{figure*}[t]
\centering
{\scriptsize
\begin{tikzpicture}[
  >=Stealth,
  node distance=6mm and 4mm,
  startend/.style={circle,draw,fill=orange!40,minimum size=6mm,inner sep=0pt},
  block/.style={rectangle,draw,rounded corners=2pt,fill=cyan!20,
                minimum width=16mm, minimum height=7mm, align=center},
  diamondshape/.style={draw,fill=cyan!20,diamond,aspect=2,
                       minimum width=14mm,minimum height=14mm,inner sep=0pt},
  dtext/.style={align=center,inner sep=0pt},
  line/.style={-{Stealth[length=0.9mm,width=0.8mm]},line width=0.5pt}
]

\node[startend] (start) {Start};
\node[block, right=7mm of start] (chooseT) {\shortstack{Choose $\pi_T$\\from $G_T$}};
\node[block, right=5mm of chooseT] (chooseA) {\shortstack{Choose $\pi_A$\\from $G_A$}};
\node[block, right=5mm of chooseA] (incOTP) {Incremental\\OTP (Alg.~\ref{alg:incremental-otp})};

\node[diamondshape, right=5mm of incOTP] (pruneGA) {};
\node[dtext] at (pruneGA.center) {\shortstack{Can\\prune\\$\pi_A$?}};

\node[diamondshape, right=5mm of pruneGA] (moreA) {};
\node[dtext] at (moreA.center) {\shortstack{More\\paths $\pi_A'$\\left?}};

\node[diamondshape, right=5mm of moreA] (pruneGT) {};
\node[dtext] at (pruneGT.center) {\shortstack{Can\\prune\\$\pi_T$?}};

\node[diamondshape, right=5mm of pruneGT] (moreT) {};
\node[dtext] at (moreT.center) {\shortstack{More\\paths $\pi_T'$\\left?}};

\node[startend, right=7mm of moreT] (end) {End};

\draw[line] (start) -- node[above,yshift=0.2mm]{$G_T$} node[below]{\(G_A\)} (chooseT);
\draw[line] (chooseT) -- node[above]{$\pi_T$} node[below]{\(\His\)} (chooseA);
\draw[line] (chooseA) -- node[above]{$\pi_A$} node[below]{\(\His\)} (incOTP);
\draw[line] (incOTP) -- node[above]{$\R$} (pruneGA);
\draw[line] (pruneGA) -- node[above]{No} (moreA);
\draw[line] (moreA) -- node[above]{No} (pruneGT);
\draw[line] (pruneGT) -- node[above]{No} (moreT);
\draw[line] (moreT) -- node[above]{No}  (end);

\node[below=0mm of pruneGA.south, xshift=2.5mm] {Yes};
\node[below=0mm of moreA.south, xshift=2.5mm] {Yes};
\node[below=0mm of pruneGT.south, xshift=2.5mm] {Yes};
\node[below=0mm of moreT.south, xshift=2.5mm] {Yes};
\draw[line] (pruneGA.south) |- ++(0,-4mm) -| node[pos=0.25,below]{} (chooseA.south);
\draw[line] (moreA.south) |- ++(0,-4mm) -| node[pos=0.25,below]{} (chooseA.south);
\draw[line] (pruneGT.south) |- ++(0,-6mm) -| node[pos=0.25,below]{} (chooseT.south);
\draw[line] (moreT.south) |- ++(0,-6mm) -| node[pos=0.25,below]{} (chooseT.south);

\draw[line, dashed] (incOTP.north) |- ++(0,4mm) -| node[pos=0.25,above]{} (chooseT.north);

\node[above=0.5mm of incOTP.north, xshift=2.5mm] {\(\His'\)};

\begin{scope}[on background layer]


\coordinate (extraTop) at ($(chooseA.north)+(0,8mm)$);
\coordinate (extraTopInner) at ($(chooseA.north)+(0,5mm)$);

\node[
  fill=gray!30,
  rounded corners,
  fit=(chooseA) (incOTP) (pruneGA) (moreA) (chooseT) (pruneGT) (moreT) (extraTop),
  inner sep=7mm,
  inner ysep=8mm,   
  label={[yshift=-5mm]90:\textbf{Outer BnB} (Alg.~\ref{alg:nested})}
] (outerbg) {};

\node[
  fill=gray!10,
  rounded corners,
  fit=(chooseA) (incOTP) (pruneGA) (moreA) (extraTopInner),
  inner sep=5mm,
  inner ysep=5mm,   
  label={[yshift=-5mm]90:\textbf{Inner BnB} (Alg.~\ref{alg:assistance-otp})}
] (innerbg) {};

\end{scope}

\end{tikzpicture}
}
\caption{Illustration of the algorithmic framework. 
    Solid lines denote data flow, dashed lines denote data structure updates.}
    \label{fig:alg-overview}
    \vspace{-5mm}
\end{figure*}

In this section we introduce our nested BnB algorithm for solving the \textsc{JointTAP} problem.
We begin by introducing  an upper bound for the outer BnB (Sec.~\ref{subsec:jointtap-ub}). Specifically, a bound on the maximum reward achievable by any assistance path in~\(G_A\) given a specific task path~\(\pi_T\). 
We then continue to describe our Nested BnB instantiation, along with additional optimizations (Sec.~\ref{subsec:jointtap-improv}). We then describe how the bound and the optimizations are used in our algorithm (Sec.~\ref{subsec:jointtap-explained}).

\subsection{A Flow-based Upper Bound}
\label{subsec:jointtap-ub}

As we will explain shortly (Sec.~\ref{subsec:jointtap-improv}), in the outer BnB it will be useful to bound the reward obtained by any extension of a task path~\(\pi_T\).
Thus, we denote~$\UB_{\text{joint}}(\bar{v}_T)$ to be an upper bound on the reward that can be obtained when~\(\taskR\) starts from a given task vertex \(\bar{v}_T \in V_T\), and \(\assistR\) may begin from any vertex in \(V_A\).

To compute $\UB_{\text{joint}}(\cdot)$, we introduce an equivalent formulation of the \textsc{JointTAP} problem given as an Integer Program (IP) encoding a network flow problem. Both the IP's objective and the fact that we solve the corresponding (relaxed) Linear Program (LP)  will allow us to compute~$\UB_{\text{joint}}(\cdot)$. 

First, we introduce the \emph{joint graph}~\(G^\times(G_A, G_T, \bar{v}_T):= (V^\times, E^\times)\) which simultaneously encodes transitions of \taskR and \assistR. Roughly speaking, transitions in the joint graph correspond to 
motions of~\taskR along edges of $E_T$, 
and motions of~\assistR along connected paths in $E_A$.
Specifically, we first define $V^\times_v:= V_T \times V_A$, and we use it to define the set of vertices as
$V^\times:= V^\times_v \cup \{s,t\}$.
Namely, $V^\times$ contains all pairs of vertices from $V_T$ and $V_A$ as well as two special vertices $s$ and $t$.
The set of edges is defined as 
$E^\times:=E^\times_e \cup E^\times_s \cup E^\times_t$
where
$\left((v_T, v_A),(\tilde{v_T}, \tilde{v_A})\right) \in E^\times_e$ if 
(i)~$(v_T, \tilde{v_T}) \in E_T$ and
(ii)~$v_A$ and $\tilde{v_A}$ lie in the same connected component of~$G_A$.
In addition, 
$E^\times_s:= \{(s,p)~\vert~p\in \{ \bar{v}_T\} \times V_A \}$ 
and 
$E^\times_t:=\{(p,t)~\vert~p \in  V^\times_v\}$.
Namely, the set of edges~$E^\times_e$ connect pairs of vertices whenever $\taskR$ can transition along an edge in $E_T$ and $\assistR$ can move to a reachable vertex.
The set of edges $E^\times_s$ connect $s$ to every vertex such that \taskR is located in $\bar{v}_T$ (regardless of the location of $\assistR$) 
while the set of edges $E^\times_t$ connect every vertex in $V^\times_v$ to $t$. 

We now define a LP that encodes a max-flow problem between $s$ and $t$ in a joint graph $G^\times( G_A, G_T, \bar{v}_T)$:
\begin{align}
\max \quad &
    \sum_{(p,\tilde{p}) \in E^\times_e}  
    \ell(v_T,\tilde{v_T}) \cdot \bigl(\mathcal{A}(p)+\mathcal{A}(\tilde{p})\bigr) \cdot f(p,\tilde{p}), 
    \label{eq:opt}\\
\text{s.t.} \quad
& \sum_{(p,\tilde{p}) \in E^\times_e} f(p,\tilde{p}) \cdot \ell(v_T,\tilde{v_T}) \leq 1,  
    \label{eq:lp-ellt}\\
& \sum_{(p,\tilde{p}) \in E^\times_e} f(p,\tilde{p}) \cdot \delta'(v_A,\tilde{v_A}) \leq 1,
    \label{eq:lp-ella} \\  
&  \sum_{(\tilde{p},p) \in E^\times_e} f(\tilde{p},p)=
   \sum_{(p,\tilde{p}) \in E^\times_e} f(p,\tilde{p}),\; \forall p \in V^\times_v \label{eq:lp-flow-a}, \\
&  \sum_{(p,t) \in E^\times_t} f(p,t) =\sum_{(s,p) \in E^\times_s} f(s,p) =1 \label{eq:lp-flow-c}.
\end{align}

Each variable \(f(p,\tilde{p})\) in the LP denotes a unit flow from vertex \(p = (v_T, v_A)\) to vertex \(\tilde{p} = (\tilde{v_T}, \tilde{v_A})\). 
Constraints~\eqref{eq:lp-ellt} and~\eqref{eq:lp-ella} ensure that the total time taken by $\taskR$ and $\assistR$ does not exceed the time limit, respectively.
Constraints~\eqref{eq:lp-flow-a}-\eqref{eq:lp-flow-c} enforces standard flow conservation for every vertex (i.e., in-flow equals out-flow except for $s$ and $t$ have only a unit of outgoing and incoming flow, respectively).

To understand the optimization (Eq.~\eqref{eq:opt}) of this max-flow problem between, 
recall that each edge~\((p, \tilde{p})\) corresponds to a simultaneous transition of \taskR  and \assistR and contributes a reward equal to \(0.5 \cdot \ell(v_T, \tilde{v_T})\) for each of the task vertices~\(v_T\) and~\(\tilde{v_T}\), if indeed assistance is supplied by the corresponding assistance vertices~\(v_A\) and~\(\tilde{v_A}\), respectively. 
To compute the total reward of a given flow~\(f(\cdot, \cdot)\), we multiply the flow value \(f(p, \tilde{p})\) by the reward associated with that edge \((p, \tilde{p})\). Summing these contributions over all edges gives the total reward induced by the flow. 
By maximizing this sum (in Eq.~\eqref{eq:opt} we multiply this sum by two; this does not change the maximization), we obtain the best reward possible in~\(G_\times\).

Instead of solving an IP, ensuring that there is one path connecting $s$ and~$t$, we solve this LP as a fractional flow problem, allowing paths to split and recombine freely. This removes constraints enforcing a single assistance path which permits \assistR to move fractionally and to provide full assistance over all shared edge durations. Consequently, we obtain an admissible upper bound on the true reward and this bound can be computed in polynomial time.

\subsection{Nested BnB Framework and Optimizations}
\label{subsec:jointtap-improv}

\subsubsection*{Algorithm}

As stated earlier, our approach solves \textsc{JointTAP} using a nested BnB structure. The outer BnB explores partial task paths in the task graph~\(G_T\), and at each node (which corresponds to a task prefix \(\pi_T\)), it launches an inner BnB to search for an optimal assistance path and timing profile in assistance graph~\(G_A\), using \textsc{AssistanceOTP} solver (suggested in~\cite{bloch2024offlinetaskassistanceplanning}). 
To avoid unnecessary computation, we use the upper bound~\(\UB_{\text{joint}}(v_T)\) introduced in Sec.~\ref{subsec:jointtap-ub} to decide whether to launch an inner BnB at all. If this bound indicates that the current branch can't improve the best reward found so far, the current task prefix can be safely pruned without further exploration. Moreover, the outer BnB passes into each inner BnB the best reward seen so far. This enables the nested BnB to prune the inner BnB if it cannot improve the best reward seen across all outer nodes and also ensures that the inner BnB explores only branches that may improve the global best reward.

We now provide a description of our nested BnB solver for \textsc{JointTAP} (Algs.~\ref{alg:nested},\ref{alg:assistance-otp} and Fig.~\ref{fig:alg-overview}).
We start with a general description, omitting the optimization which is highlighted in \textcolor{teal}{teal} (will be explained shortly, can be ignored for now).

The outer BnB (Alg.~\ref{alg:nested}) performs a branch-and-bound search over~\(G_T\). It initializes a FIFO queue~\(Q_T\) of task paths (initialized to the start vertex~$v_0^T$), and a variable~\(\R_{\text{max}}\) to track the best reward (initialized to~0) (Line~\ref{line:jointtap-inita}). 
As long as there are paths in the queue~$Q_T$  (Line~\ref{line:jointtap-whilequeue}), the algorithm pops a path~\(\pi_T\) (Line~\ref{line:jointtap-popqueue}) and extends it to all neighbors~\(v_{k+1}^T \in V_T\) (Line~\ref{line:jointtap-foreach}) to obtain a new path~\(\pi_T'\) (Line~\ref{line:jointtap-newpath}).
Subsequently, the inner BnB (Line~\ref{line:jointtap-assistanceotp} and Alg.~\ref{alg:assistance-otp}) is called to obtain the best reward~\(\R\) for the fixed path~\(\pi_T'\) and for every path of~\assistR. 

\begin{algorithm}[t!]
\caption{Nested BnB \textsc{JointTAP} Solver}\label{alg:nested}
\hspace*{\algorithmicindent} 
\textbf{Input:} 
Assistance $\A$;
graphs $G_A,G_T$; 
start vertices $v_0^A,v_0^T$\\
\hspace*{\algorithmicindent} 
\textbf{Output:} Optimal reward~\(\R_{\text{max}}\)
\begin{algorithmic}[1]

\State \( Q_T\gets \{(\langle v_0^T\rangle, \textcolor{teal}{\emptyset})\} \); \quad \(\R_{\text{max}} \gets 0\)  \label{line:jointtap-inita}

    \While{\(Q_T \ne \emptyset\)} \label{line:jointtap-whilequeue}

        \State \((\pi_T=\langle v_0^T,\ldots,v_k^T\rangle,\textcolor{teal}{\hists_{\pi_T}}) \gets Q_T.\texttt{pop}()\) \label{line:jointtap-popqueue}
        \For{\textbf{each} $v_{k+1}^T$ s.t. $(v_k^T, v_{k+1}^T) \in E_T$} \label{line:jointtap-foreach} \Comment{Branch}
            \State $\pi_T' \gets \langle v_0^T,\ldots,v_{k+1}^T\rangle$ \label{line:jointtap-newpath}

            \State \textcolor{teal}{\(\I \gets \texttt{ComputeIntervals}(v_{k+1}^T,\A)\)  \label{line:jointtap-comupteintervals}} 
            \State \(\R,\textcolor{teal}{\hists_{\pi_T'}}\gets \texttt{IB}(\pi_T,G_A, v_0^A,\R_\text{max},\textcolor{teal}{\hists_{\pi_T}},\textcolor{teal}{\I})\) \label{line:jointtap-assistanceotp} \Comment{Alg.~\ref{alg:assistance-otp}}
            \If{\(\R + \UB_\text{joint}(v_{k+1}^T)\le \R_\text{max}\)}  \label{line:jointtap-checkprune} 
                 \textbf{continue}\label{line:jointtap-prune} \Comment{Prune}
            \EndIf
        \State \(\R_\text{max} \gets \max\{\R_\text{max},\R\}\) \label{line:jointtap-checklocal}


            \State \(Q_T\).\texttt{push}(\((\pi_T',\textcolor{teal}{\hists_{\pi_T'}})\)) \label{line:jointtap-push}
        \EndFor
    \EndWhile
    
\State \Return \(\R_\text{max}\) \label{line:jointtap-ret}

\end{algorithmic}
\end{algorithm}

\begin{algorithm}[t!]
\caption{\texttt{IB} (Inner BnB)}\label{alg:assistance-otp}
\hspace*{\algorithmicindent} 
\textbf{Input:} 
Task path $\pi_T$; 
Assistance graph $G_A$; \\
{\color{white}.\hspace{12mm}}
start vertex $v_0^A$;
current reward \(\R_\text{curr}\);\\
%
%
{\color{white}.\hspace{12mm}}
\textcolor{teal}{
list of histories \(\hists_{\pi_T'}\);
intervals \(\I\)  of last vertex in~$\pi_T$}\\
\hspace*{\algorithmicindent} 
\textbf{Output:} Optimal reward~\(\R_{\text{max}}\); 
\textcolor{teal}{
list of histories~\(\hists_{\pi_T'}\)}
\begin{algorithmic}[1]

\State \(Q_A \gets \{\langle v_0^A \rangle\}\) ;\quad \(\R_{\text{max}} \gets \R_{\text{curr}}\) 
 ; \quad \textcolor{teal}{\(\hists_{\pi_T'}\gets \emptyset\)} \label{line:assistanceotp-inita} 

%

\While{\(Q_A \ne \emptyset\)} \label{line:assistanceotp-while}
    
    \State \(\pi_A =\langle v_0^A,\ldots,v_k^A\rangle\gets Q_A.\texttt{pop}()\) \label{line:assistanceotp-pop}
    \For{\textbf{each} $v_{k+1}^A$ s.t. $(v_k^A, v_{k+1}^A) \in E_A$} \label{line:assistanceotp-foreach} \Comment{Branch}
        \State $\pi_A' \gets \langle v_0^A,\ldots,v_{k+1}^A\rangle$ \label{line:assistanceotp-newpath} 
        
\State $\R \gets \texttt{OTP}(\pi_A', \pi_T, \A)$  
    \label{line:assistanceotp-otp}
    \Comment{See Bloch and Salzman~\cite{bloch2024offlinetaskassistanceplanning}}

        \State \textcolor{teal}{\(\R,\hists_{\pi_T'}(\pi_A')\gets \texttt{IncOTP}(\hists_{\pi_T}(\pi_A'),\I,\pi_A')\) \Comment{Alg.\ref{alg:incremental-otp}} \label{line:assistanceotp-incotp} }

        \If {\(\UB_A(\pi_A') \le \textcolor{violet}{\textcolor{black}{\R_\text{max}}}\)} \label{line:assistanceotp-checkprune}     
             \textbf{continue}\label{line:assistanceotp-prune} \Comment{Prune}
        \EndIf

        \State \(\R_\text{max} \gets \max\{\R_\text{max},\R\}\) \label{line:assistanceotp-maxr}

        

        \State \(Q_A\).\texttt{push}(\(\pi_A'\)) \label{line:assistanceotp-push}
    \EndFor

\EndWhile
\State \Return \(\R_\text{max}, \textcolor{teal}{\hists_{\pi_T'}}\) 
\label{line:assistanceotp-ret}
\end{algorithmic}
\end{algorithm}

Once the reward is computed, we compute an upper bound~\(\UB_\text{joint}(\cdot)\) on the reward of any subpath of~\(\pi_T'\) (Line~\ref{line:jointtap-checkprune} and Sec.~\ref{subsec:jointtap-ub}). This is used to prune the subtree extending~\(\UB_\text{joint}(\cdot)\) in case the best reward~\(\R_{\text{max}}\) can't be improved (Line~\ref{line:jointtap-prune}).
Otherwise, the maximum reward is updated (Line~\ref{line:jointtap-checklocal}).
The new path~\(\pi_T'\) is added to~\(Q_T\) for further exploration (Line~\ref{line:jointtap-push}). Once the task search terminates, the best reward~\(\R_{\text{max}}\) is returned (Line~\ref{line:jointtap-ret}).

The inner BnB (Alg.~\ref{alg:assistance-otp}) performs a branch-and-bound search over the assistance graph~\(G_A\).
It initializes a queue~\(Q_A\) of assistance paths (initialized to the start vertex~$V_0^A$), and a variable~\(\R_{\text{max}}\) to track the best reward (initialized to the current-best reward~$\R_{\text{curr}}$ of the outer BnB) (Line~\ref{line:assistanceotp-inita}). 
As long as there are paths in the queue~$Q_A$  (Line~\ref{line:assistanceotp-while}), the algorithm pops a path~\(\pi_A\) (Line~\ref{line:assistanceotp-pop})  and extends it to all neighbors~\(v_{k+1}^A\) in $G_A$ (Line~\ref{line:assistanceotp-foreach}) to obtain a new path~\(\pi_A'\) (Line~\ref{line:assistanceotp-newpath}).

The reward~\(\R\) for~\(\pi_A'\) and~\(\pi_T\) is computed using an OTP solver  (Line~\ref{line:assistanceotp-otp}) and if a bound on the reward obtainable from the given assistance path~\(\pi'_A\) (Line~\ref{line:assistanceotp-checkprune}) cannot improve upon the current-maximal reward~$R_{\max}$, the node is pruned (Line~\ref{line:assistanceotp-prune}). Here, we use the OTP solver and bound~$\UB_A(\cdot)$ introduced by Bloch and Salzman~\cite{bloch2024offlinetaskassistanceplanning}.
Otherwise, the inner maximum reward is updated (Line~\ref{line:assistanceotp-maxr}) and the new path~\(\pi_A'\) is added to~\(Q_A\) (Line~\ref{line:assistanceotp-push}). Once the inner BnB terminates, the best reward~\(\R_{\text{max}}\) is returned to the outer BnB (Line~\ref{line:assistanceotp-ret}).





\subsubsection*{Using incremental OTP} We suggest the following optimization (highlighted in \textcolor{teal}{teal} in Alg.~\ref{alg:nested} and Alg.~\ref{alg:assistance-otp}):
As the outer BnB explores different task path prefixes, each inner BnB differs from its parent by the addition of a single task vertex. To exploit this, we harness our incremental OTP solver (Sec.~\ref{sec:inc-otp}). 
{Specifically, in the outer BnB, we save for each branch~\(\pi_T\) a list of histories~\(\hists_{\pi_T}\). Each history in this list, denoted~\(\hists_{\pi_T}(\pi_A)\), represents the history of all intervals~\(\I_{\pi_T}^\A\) created by~\(\pi_T\) for the vertices in the assistance path~\(\pi_A\). Thus, we store in the queue~\(Q_T\), along with~\(\pi_T\), the corresponding list of histories (Alg.~\ref{alg:nested}, Line~\ref{line:jointtap-popqueue} and~\ref{line:jointtap-push}). Each time we invoke the inner BnB for a child~\(\pi_T'\) of~\(\pi_T\), we use the list of histories~\(\hists_{\pi_T}\) of~\(\pi_T\) to consider only the new intervals created by the last vertex in~\(\pi_T'\). This is done using function \texttt{ComputeIntervals} (Line~\ref{line:jointtap-comupteintervals}. Description omitted).  We then compute the reward of the optimal timing profile for~\(\pi_A\) and~\(\pi_T'\) based on the prior calculation for~\(\pi_A\) and~\(\pi_T\), and use the updated history 
to create the new list of histories~\(\hists_{\pi_T'}\) for~\(\pi_T'\) (Alg.~\ref{alg:assistance-otp}, Line~\ref{line:assistanceotp-incotp}; replaces Line~\ref{line:assistanceotp-otp}).}

\label{subsec:jointtap-explained}

\section{Empirical Evaluation}
\label{sec:experiments}

\begin{figure}[t!]    
  \centering
 \subfloat[]{\includegraphics[height = 2.2cm, trim={1.5cm 1.cm 1cm 1.cm}, clip]{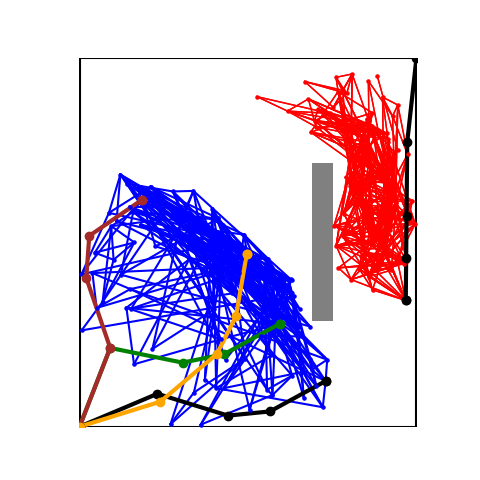}\label{fig:one_obs}}
  \hspace{0.7cm} 
  \subfloat[]{\includegraphics[height = 2.2cm, trim={1.5cm 1.cm 1.5cm 1.cm}, clip]{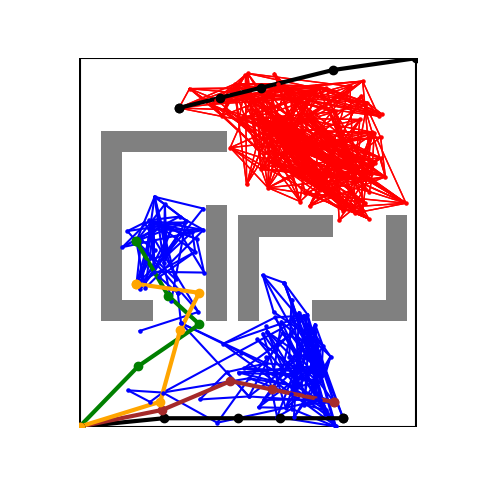}\label{fig:two_rooms}}
  \hspace{0.7cm} 
  \subfloat[]{\includegraphics[height=2.2cm, trim={1cm 1.cm 1.5cm 1.cm}, clip]{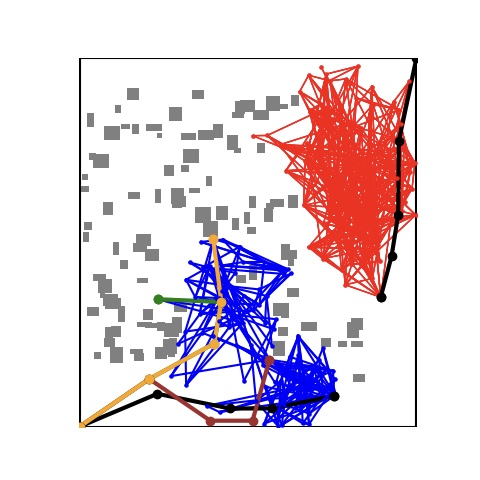}\label{fig:rand_obs}} \\
    \hfill
  \subfloat[]{\includegraphics[height=2.1cm]{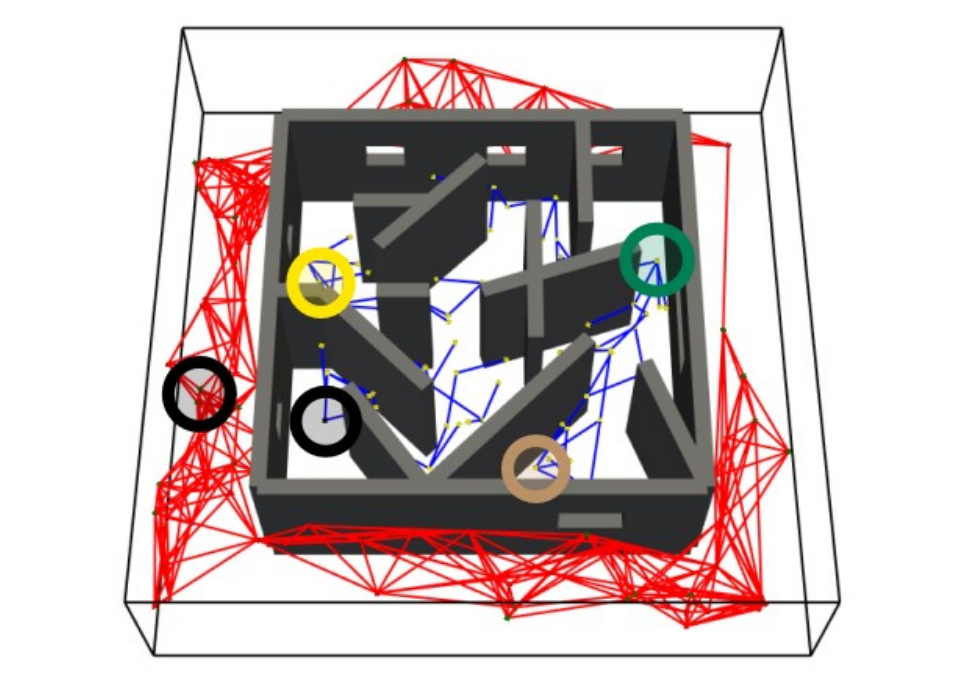}\label{fig:open_space}}
  \hfill
  \subfloat[]{\includegraphics[height = 2.1cm]{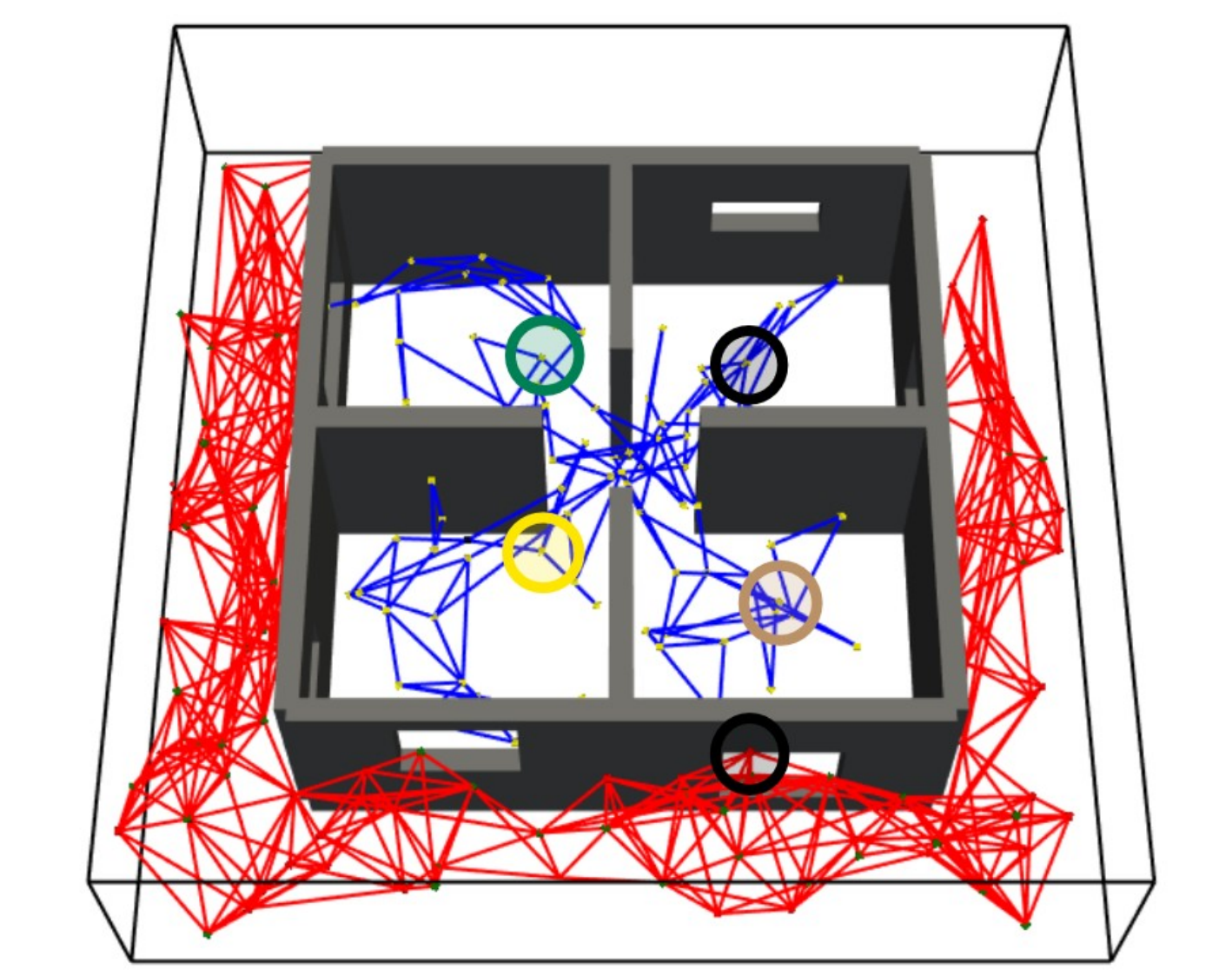}\label{fig:four_rooms}}%
  \hfill
  \subfloat[]{\includegraphics[height = 2.1cm]{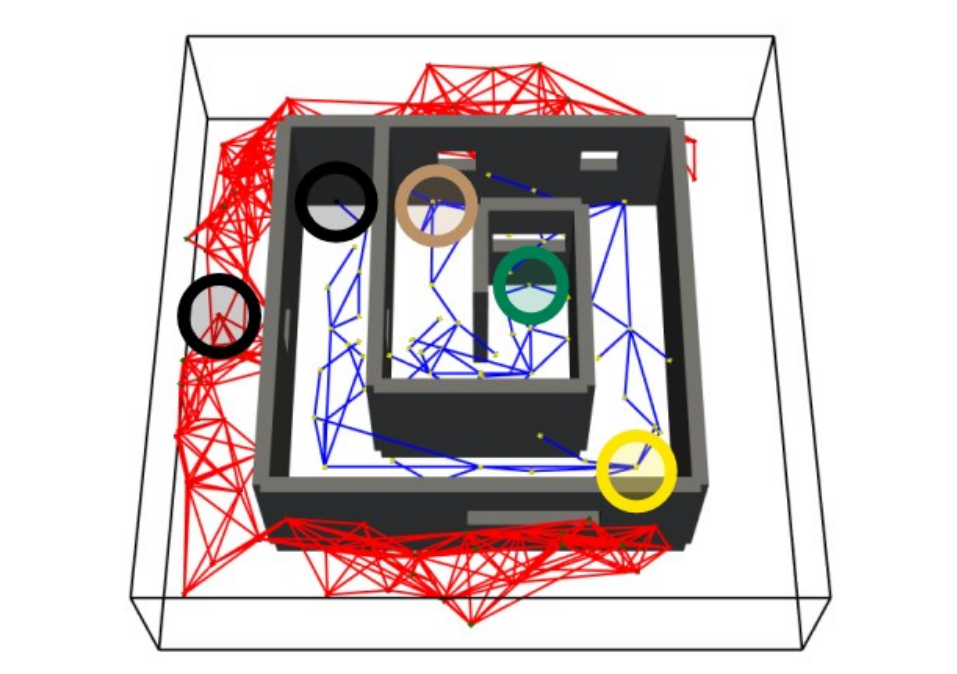}\label{fig:spiral}}%

  \caption{%
  Visualization of representative simulated environments 
  for planar manipulator (a)-(c) and drones (d)-(f).
  Each consists 
  one instance of \(G_A\) (red) and \(G_T\) (blue),
  the robots \taskR,\assistR (black at their start vertices) and obstacles (gray). 
  %
  The task of \taskR is to move from start configuration (black) to its target configuration (green) while passing through waypoints (yellow, then brown) and staying on $G_T$.
  This should be done while maximizing the overall time its end effector is within Line of Sight of \assistR.}  
  
  \label{fig:exp-envs}
\end{figure}

We consider the \textsc{JointTap} problem of LOS-maintenance (Sec.~\ref{sec:intro}  and  Fig.~\ref{fig:motivating-example}) for both 
(i)~a toy, 2D setting where $\assistR$ and $\taskR$ are modeled as {4-DOF} planar manipulators and
(ii)~a simulated 3D setting where $\assistR$ and $\taskR$ are modeled as Crazyflie~2.1+ drones---a lightweight, open-source nano-quadrotor platform.
See (Fig.~\ref{fig:exp-envs}) for representative simulated environments.
We also conducted real-world evaluation in the lab using two Crazyflie~2.1+ drones (see accompanying video).
For each simulated scenario, we construct {ten pairs of} roadmaps, \(G_A\) and \(G_T\) 
where 
\(G_A\) is constructed using the \textsf{RRG} algorithm~\cite{KF11}, 
while
\(G_T\) is generated by specifying a sequence of predefined waypoints (representing the general task need to be completed) and running the  \textsf{RRG} algorithm  between consecutive waypoints.

We compare three different algorithms \textsf{DFS}, \textsf{BnB} and \textsf{BnB-Inc}.
\textsf{DFS} is a baseline algorithm that exhaustively enumerates all possible task paths in $G_T$ using a depth-first search (DFS) approach, halting when the path length exceeds~1.
Reward for each path in  $G_T$ is computed using the state-of-the-art \textsc{AssistanceOTP} algorithm~\cite{bloch2024offlinetaskassistanceplanning}. Importantly, \textsf{DFS} is optimal and  equivalent to our Nested BnB algorithm when using a trivial upper bound of~1 and omitting all optimizations. 
\textsf{BnB} and \textsf{BnB-Inc} are our proposed approach (Alg.~\ref{alg:nested}) with and without the optimizations, respectively.

Algorithms were implemented in C\raise.08ex\hbox{\tt ++}\xspace\footnote{\href{https://github.com/CRL-Technion/JointTAP}{github.com/CRL-Technion/JointTAP}}, and experiments were conducted on a Dell Inspiron 5410 laptop with 16\,GB of RAM. Each algorithm was given a timeout of one hour.

\begin{figure}[t]
    \centering

    \begin{tikzpicture}
        \path[draw=none] (0,0) -- (1,0);
        \begin{axis}[
            hide axis,
            xmin=0, xmax=1,
            ymin=0, ymax=1,
            legend style={
                at={(0.5,1.0)},
                anchor=south,
                legend columns=4,
                column sep=6pt,
                /tikz/every even column/.append style={column sep=6pt},
                font=\footnotesize,
                draw=none
            }
        ]
        \addlegendimage{mark=x, dashed, semithick, black, mark size=1.0}
        \addlegendentry{Reward}
        \addlegendimage{mark=square*, semithick, green!60!black, mark size=1.0}
        \addlegendentry{\textsf{DFS}}
        \addlegendimage{mark=triangle*, semithick, blue, mark size=1.0}
        \addlegendentry{\textsf{BnB}}
        \addlegendimage{mark=*, semithick, orange, mark size=1.0}
        \addlegendentry{\textsf{BnB-Inc}}
        \end{axis}
    \end{tikzpicture}

    \vspace{-16em}

    \begin{tikzpicture}
        \node at (0,0) {
            \begin{tikzpicture}
                \begin{axis}[
                    width=3.7cm,
                    height=3cm,
                    xlabel={Graph Size [Vertices]},
                    ylabel={Runtime [min]},
                    tick align=outside,
                    tick pos=left,
                    xtick={0,20,40,60,80,100},
                    ytick={0,10,20,30,40,50,60},
                    ymin=0, ymax=60,
                    xmin=0, xmax=100,
                    tick label style={font=\scriptsize},
                    label style={font=\scriptsize}
                ]
                \addplot+[mark=triangle*, semithick, blue, mark size=1.0] coordinates {
                    (10,0.058/60) (20,0.084/60) (30,4.445/60)
                    (40,31.746/60) (50,83.517/60) (60,124.099/60)
                    (70,385.243/60) (80,555.782/60)
                    (90,556.854/60) (100,1388.71/60)
                };
                \addplot+[mark=*, semithick, orange, mark size=1.0] coordinates {
                    (10,0.002/60) (20,0.015/60) (30,0.323/60)
                    (40,3.713/60) (50,14.271/60) (60,30.013/60)
                    (70,86.929/60) (80,123.948/60)
                    (90,124.585/60) (100,410.26/60)
                };
                \addplot+[mark=square*, semithick, green!60!black, mark size=1.0] coordinates {
                    (10,0.018/60) (20,0.544/60) (30,104.047/60)
                    (40,1820.506/60)
                };
                \end{axis}
                            \begin{axis}[
                width=3.7cm,
                height=3cm,
                xlabel={},
                tick align=outside,
                tick pos=right,
                axis y line*=right,
                axis x line=none,
                ylabel={},
                ytick={0,0.2,0.4,0.6,0.8,1.0},
                ymin=0, ymax=1,
                xmin=0, xmax=100,
                tick label style={font=\scriptsize},
                label style={font=\scriptsize}
            ]
            \addplot+[mark=x, dashed, semithick, black, mark size=1.0] coordinates {
                (10,0) (20,0) (30,0.065078) (40,0.10813)
                (50,0.134321) (60,0.177376) (70,0.239986)
                (80,0.307641) (90,0.307641) (100,0.33727)
            };
            \end{axis}

            \end{tikzpicture}
        };
        \node at (4.7,0) {
            \begin{tikzpicture}
                \begin{axis}[
                    width=3.7cm,
                    height=3cm,
                    xlabel={Graph Size [Vertices]},
                    ylabel={},
                    tick align=outside,
                    tick pos=left,
                    xtick={0,20,40,60,80,100},
                    ytick={0,10,20,30,40,50,60},
                    ymin=0, ymax=60,
                    xmin=0, xmax=100,
                    tick label style={font=\scriptsize},
                    label style={font=\scriptsize}
                ]
                            \addplot+[mark=triangle*, semithick, blue, mark size=1.0] coordinates {
                (10,0.039/60) (20,0.181/60) (30,0.225/60)
                (40,4.228/60) (50,30.468/60) (60,56.144/60)
                (70,198.237/60) (80,1562.47/60)
                (90,1562.99/60)
            };
            \addplot+[mark=*, semithick, orange, mark size=1.0] coordinates {
                (10,0.014/60) (20,0.089/60) (30,0.132/60)
                (40,2.157/60) (50,8.054/60) (60,15.568/60)
                (70,43.443/60) (80,530.587/60)
                (90,530.882/60)
            };
            \addplot+[mark=square*, semithick, green!60!black, mark size=1.0] coordinates {
                (10,0.154/60) (20,0.741/60) (30,2.559/60)
                (40,16.052/60) (50,191.033/60)
            };

                \end{axis}
                            \begin{axis}[
                width=3.7cm,
                height=3cm,
                xlabel={},
                tick align=outside,
                tick pos=right,
                axis y line*=right,
                axis x line=none,
                ylabel={Quality [Reward]},
                ytick={0,0.2,0.4,0.6,0.8,1.0},
                ymin=0, ymax=1,
                xmin=0, xmax=100,
                tick label style={font=\scriptsize},
                label style={font=\scriptsize}
            ]
            \addplot+[mark=x, dashed, semithick, black, mark size=1.0] coordinates {
                (10,0.0615841) (20,0.0687319) (30,0.0687319) (40,0.0833771)
                (50,0.153757) (60,0.294842) (70,0.454463)
                (80,0.622156) (90,0.622156)
            };
            \end{axis}

            \end{tikzpicture}

        };
            \node at (0,-1.7) {\centering \footnotesize (a) Results for Fig.~\ref{fig:one_obs} \par};
            \node at (4.7,-1.7) {\centering \footnotesize (b) Results for Fig.~\ref{fig:two_rooms} \par};

    \end{tikzpicture}

    \begin{tikzpicture}
        \node at (0,0) {
        \begin{tikzpicture}
            \begin{axis}[
                width=3.7cm,
                height=3cm,
                xlabel={Graph Size [Vertices]},
                ylabel={Runtime [min]},
                tick align=outside,
                tick pos=left,
                xtick={0,20,40,60,80,100},
                ytick={0,10,20,30,40,50,60},
                ymin=0, ymax=60,
                xmin=0, xmax=100,
                tick label style={font=\scriptsize},
                label style={font=\scriptsize}
            ]

        \addplot+[mark=triangle*, semithick, blue, mark size=1.0] coordinates {
            (10,0.213/60) (20,3.174/60) (30,19.597/60)
            (40,103.177/60) (50,165.913/60) (60,485.502/60)
            (70,1131.08/60) (80,2308.89/60)
        };
        \addplot+[mark=*, semithick, orange, mark size=1.0] coordinates {
            (10,0.025/60) (20,0.507/60) (30,3.818/60)
            (40,13.969/60) (50,30.693/60) (60,108.759/60)
            (70,305.181/60) (80,687.526/60)
        };
        \addplot+[mark=square*, semithick, green!60!black, mark size=1.0] coordinates {
            (10,0.216/60) (20,5.332/60) (30,164.092/60)
            (40,725.876/60)
        };

            \end{axis}
                        \begin{axis}[
            width=3.7cm,
            height=3cm,
            xlabel={},
            tick align=outside,
            tick pos=right,
            axis y line*=right,
            axis x line=none,
            ylabel={},
            ytick={0,0.2,0.4,0.6,0.8,1.0},
            ymin=0, ymax=1,
            xmin=0, xmax=100,
            tick label style={font=\scriptsize},
            label style={font=\scriptsize}
        ]
        \addplot+[mark=x, dashed, semithick, black, mark size=1.0] coordinates {
            (10,0.120641) (20,0.180723) (30,0.252194) (40,0.376197)
            (50,0.527764) (60,0.544352) (70,0.544352) (80,0.544352)
        };

        \end{axis}

        \end{tikzpicture}

    };
            \node at (4.7,0) {
        \begin{tikzpicture}
            \begin{axis}[
                width=3.7cm,
                height=3cm,
                xlabel={Graph Size [Vertices]},
                ylabel={},
                tick align=outside,
                tick pos=left,
                xtick={0,20,40,60,80,100},
                ytick={0,10,20,30,40,50,60},
                ymin=0, ymax=60,
                xmin=0, xmax=100,
                tick label style={font=\scriptsize},
                label style={font=\scriptsize}
            ]
            \addplot+[mark=triangle*, semithick, blue, mark size=1.0] coordinates {
                (10,0.0/60) (20,0.042/60) (30,0.609/60) (40,28.132/60)
                (50,30.956/60) (60,31.008/60) (70,291.167/60)
                (80,624.878/60) (90,1498.94/60) (100,3375.37/60)
            };
            \addplot+[mark=*, semithick, orange, mark size=1.0] coordinates {
                (10,0.0/60) (20,0.018/60) (30,0.186/60) (40,16.157/60)
                (50,18.773/60) (60,18.876/60) (70,239.53/60)
                (80,516.474/60) (90,585.54/60) (100,1529.14/60)
            };
            \addplot+[mark=square*, semithick, green!60!black, mark size=1.0] coordinates {
                (10,0.031/60) (20,0.628/60) (30,27.248/60)
                (40,686.906/60)
            };

            \end{axis}
                        \begin{axis}[
            width=3.7cm,
            height=3cm,
            xlabel={},
            tick align=outside,
            tick pos=right,
            axis y line*=right,
            axis x line=none,
            ylabel={Quality [Reward]},
            ytick={0,0.2,0.4,0.6,0.8,1.0},
            ymin=0, ymax=1,
            xmin=0, xmax=100,
            tick label style={font=\scriptsize},
            label style={font=\scriptsize}
        ]
            \addplot+[mark=x, dashed, semithick, black, mark size=1.0] coordinates {
                (10,0.0) (20,0.0793131) (30,0.128935) (40,0.18858)
                (50,0.29382) (60,0.29382) (70,0.310282) (80,0.310282)
                (90,0.310282) (100,0.375541)
            };

        \end{axis}

        \end{tikzpicture}
    };
        \node at (0,-1.7) {\centering \footnotesize (c) Results for Fig.~\ref{fig:rand_obs}\par};
        \node at (4.7,-1.7) {\centering \footnotesize (d) Results for Fig.~\ref{fig:open_space} \par};

    \end{tikzpicture}
    
    \begin{tikzpicture}
        \node at (0,0) {
        \begin{tikzpicture}
            \begin{axis}[
                width=3.7cm,
                height=3cm,
                xlabel={Graph Size [Vertices]},
                ylabel={Runtime [min]},
                tick align=outside,
                tick pos=left,
                xtick={0,20,40,60,80,100},
                ytick={0,10,20,30,40,50,60},
                ymin=0, ymax=60,
                xmin=0, xmax=100,
                tick label style={font=\scriptsize},
                label style={font=\scriptsize}
            ]
                            \addplot+[mark=triangle*, semithick, blue, mark size=1.0] coordinates {
                (10,0.008/60) (20,0.242/60) (30,0.654/60) (40,1.581/60) (50,3.124/60)
                (60,21.109/60) (70,83.367/60) (80,158.473/60) (90,903.35/60) (100,2709.87/60)
            };
            \addplot+[mark=*, semithick, orange, mark size=1.0] coordinates {
                (10,0.005/60) (20,0.233/60) (30,0.561/60) (40,1.365/60) (50,2.757/60)
                (60,19.812/60) (70,73.531/60) (80,108.585/60) (90,508.46/60) (100,1638.18/60)
            };
            \addplot+[mark=square*, semithick, green!60!black, mark size=1.0] coordinates {
                (10,0.011/60) (20,0.238/60) (30,6.991/60)
                (40,147.133/60) (50,689.957/60)
            };

            \end{axis}
                        \begin{axis}[
            width=3.7cm,
            height=3cm,
            xlabel={},
            tick align=outside,
            tick pos=right,
            axis y line*=right,
            axis x line=none,
            ylabel={},
            ytick={0,0.2,0.4,0.6,0.8,1.0},
            ymin=0, ymax=1,
            xmin=0, xmax=100,
            tick label style={font=\scriptsize},
            label style={font=\scriptsize}
        ]
                        \addplot+[mark=x, dashed, semithick, black, mark size=1.0] coordinates {
                (10,0.0181818) (20,0.15075) (30,0.237388)
                (40,0.377362) (50,0.42897) (60,0.524586)
                (70,0.596031) (80,0.795015) (90,0.795015) (100,0.848314)
            };

        \end{axis}

        \end{tikzpicture}
    };
        \node at (4.7,0) {
        \begin{tikzpicture}
            \begin{axis}[
                width=3.7cm,
                height=3cm,
                xlabel={Graph Size [Vertices]},
                ylabel={},
                tick align=outside,
                tick pos=left,
                xtick={0,20,40,60,80,100},
                ytick={0,10,20,30,40,50,60},
                ymin=0, ymax=60,
                xmin=0, xmax=100,
                tick label style={font=\scriptsize},
                label style={font=\scriptsize}
            ]
                                \addplot+[mark=triangle*, semithick, blue, mark size=1.0] coordinates {
                (10,0.0/60) (20,0.041/60) (30,0.478/60) (40,91.877/60)
                (50,121.641/60) (60,137.416/60) (70,1947.52/60) (80,1990.3/60)
            };
            \addplot+[mark=*, semithick, orange, mark size=1.0] coordinates {
                (10,0.0/60) (20,0.021/60) (30,0.559/60) (40,56.139/60)
                (50,73.74/60) (60,85.248/60) (70,501.529/60) (80,548.925/60) (90,1960.88/60)
            };
            \addplot+[mark=square*, semithick, green!60!black, mark size=1.0] coordinates {
                (10,0.005/60) (20,0.091/60) (30,8.779/60)
                (40,1510.94/60)
            };

            \end{axis}
                        \begin{axis}[
            width=3.7cm,
            height=3cm,
            xlabel={},
            tick align=outside,
            tick pos=right,
            axis y line*=right,
            axis x line=none,
            ylabel={Quality [Reward]},
            ytick={0,0.2,0.4,0.6,0.8,1.0},
            ymin=0, ymax=1,
            xmin=0, xmax=100,
            tick label style={font=\scriptsize},
            label style={font=\scriptsize}
        ]
                        \addplot+[mark=x, dashed, semithick, black, mark size=1.0] coordinates {
                (10,0.0) (20,0.052764) (30,0.121232) (40,0.181678)
                (50,0.34802) (60,0.34802) (70,0.514028) (80,0.514028) (90,0.54902)
            };

        \end{axis}

        \end{tikzpicture}

    };
        \node at (0,-1.7) {\centering \footnotesize (e) Results for Fig.~\ref{fig:four_rooms} \par};
        \node at (4.7,-1.7) {\centering \footnotesize (f) Results for Fig.~\ref{fig:spiral} \par};

    \end{tikzpicture}

    \caption{
    Average running time and reward (left and right $y$-axis, respectively) as a function of graph size.
    (a)-(f) correspond to environments (a)-(f) in Fig.~\ref{fig:exp-envs}.}
    \label{fig:runtimes}
\end{figure}
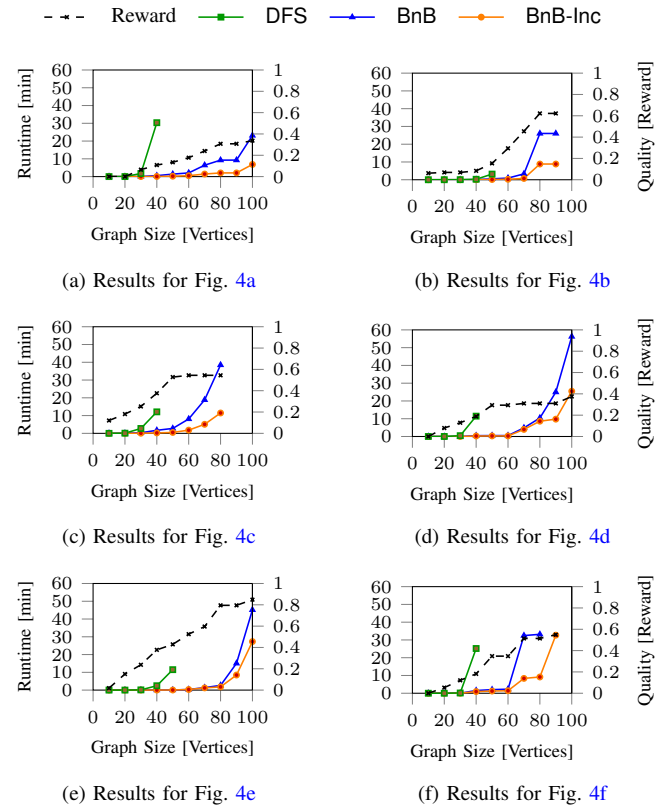

We present results for representative scenarios in Fig.~\ref{fig:runtimes}.
Results depict reward and runtime as a function of graph size (we keep $\vert G_T\vert = \vert G_A\vert$ in all experiments). 

Across all plots reward (which is identical for all algorithms as they are optimal) grows with graph size but this incurs a dramatic increase in runtime for all algorithms. When no results are provided, the timeout has been reached.

Key to the efficiency of \textsf{BnB} and \textsf{BnB-Inc} when compared to \textsf{DFS} is the the use of \(\UB_\text{joint}\).
\(\UB_\text{joint}\) both allows to prune 
paths in the outer BnB (\(G_T\)) and to obtain effective rewards that, in turn, prune paths in the inner BnB (\(G_A\))
Indeed, as can be seen in all plots, \textsf{DFS} is only able to solve instances of relatively small size and requires roughly two orders of magnitude more time than \textsf{BnB-Inc}. 

The results depict the effectiveness of the optimization. While OTP can be solved efficiently, the number of times it is required may incur  computational overhead. 
By reusing results from previous computations, we obtain a speed up of roughly $3\times$ when comparing  \textsf{BnB} with \textsf{BnB-Inc}. 
%
%

\section*{Conclusion}
In this paper, we introduced a new algorithmic framework for TAP that combines an incremental subproblem solver with a nested branch-and-bound search. The method provides optimality guarantees, and our empirical evaluation demonstrates its ability to compute coordinated plans on moderately sized instances. However, the approach incurs high computational cost on very large graphs. Future work will focus on improving scalability through stronger heuristics for this problem.
We foresee the work presented here being used in alternative problem formulations such as the online setting or for alternative cost functions (e.g., when minimizing the maximum time a path is uncovered).

\bibliographystyle{IEEEtran}
\bibliography{IEEEabrv,references}

\end{document}